\documentclass{article}

\usepackage{microtype}
\usepackage{graphicx}
\usepackage{booktabs} 

\usepackage{hyperref}

\usepackage[accepted]{icml2023}

\usepackage{amsmath}
\usepackage{amssymb}
\usepackage{mathtools}
\usepackage{amsthm}

\usepackage[capitalize,noabbrev]{cleveref}

\theoremstyle{plain}

\theoremstyle{definition}

\theoremstyle{remark}

\usepackage[textsize=tiny]{todonotes}

\newcommand{\citeg}[1]{\citep[e.g.,][]{#1}}

\usepackage{booktabs} 
\usepackage{threeparttable}
\usepackage{multirow}
\usepackage{algorithm, algorithmic}
\usepackage{float}
\usepackage{enumitem}
\usepackage[export]{adjustbox}
\usepackage{multicol}
\usepackage{subcaption}
\usepackage{wrapfig}
\usepackage{xspace}
\usepackage{soul}
\newcommand{\Algo}{\textsc{Orca}\xspace}

\usepackage{amsmath,amsfonts,bm}

\newcommand{\OT}{\text{OT}}
\newcommand{\OTDD}{\text{OTDD}}
\newcommand{\res}[2]{#1$\pm$#2}

\def\eqref#1{equation~\ref{#1}}

\def\1{\bm{1}}

\DeclareMathAlphabet{\mathsfit}{\encodingdefault}{\sfdefault}{m}{sl}
\SetMathAlphabet{\mathsfit}{bold}{\encodingdefault}{\sfdefault}{bx}{n}

\icmltitlerunning{Cross-Modal Fine-Tuning: Align then Refine}

\begin{document}

\twocolumn[
\icmltitle{Cross-Modal Fine-Tuning: Align then Refine}



\icmlsetsymbol{equal}{*}

\begin{icmlauthorlist}
\icmlauthor{Junhong Shen}{yyy,comp}
\icmlauthor{Liam Li}{comp}
\icmlauthor{Lucio M. Dery}{yyy}
\icmlauthor{Corey Staten}{comp}
\icmlauthor{Mikhail Khodak}{yyy}
\icmlauthor{Graham Neubig}{yyy}
\icmlauthor{Ameet Talwalkar}{yyy,comp}
\end{icmlauthorlist}

\icmlaffiliation{yyy}{Carnegie Mellon University}
\icmlaffiliation{comp}{Hewlett Packard Enterprise}

\icmlcorrespondingauthor{Junhong Shen}{junhongs@andrew.cmu.edu}


\vskip 0.3in
]



\printAffiliationsAndNotice{} 

\begin{abstract}
\looseness=-1
Fine-tuning large-scale pretrained models has led to tremendous progress in well-studied modalities such as vision and NLP.  
However, similar gains have not been observed in many other  modalities due to a  lack of relevant pretrained models.  In this work, we propose \Algo, a general \textit{cross-modal fine-tuning} framework that extends the applicability of a single large-scale pretrained model to diverse modalities. \Algo adapts to a target task  via an align-then-refine workflow: given the target input, \Algo first learns an embedding network  that aligns the embedded feature distribution with the pretraining modality. 
The pretrained model is then fine-tuned  on the embedded data to exploit the  knowledge shared across modalities.  Through extensive experiments, we show that \Algo obtains state-of-the-art results on 3 benchmarks 
containing over 60 datasets from 12 modalities,   
outperforming a wide range of  hand-designed, AutoML, general-purpose, and task-specific methods. We  highlight the importance of data alignment  via a series of ablation studies and demonstrate \Algo's utility in data-limited regimes.
\end{abstract}

\vspace{0.15cm}
\section{Introduction}

\begin{figure*}[t]
  \centering
  \vspace{-0.15cm}
 \includegraphics[width=0.9\textwidth]
 {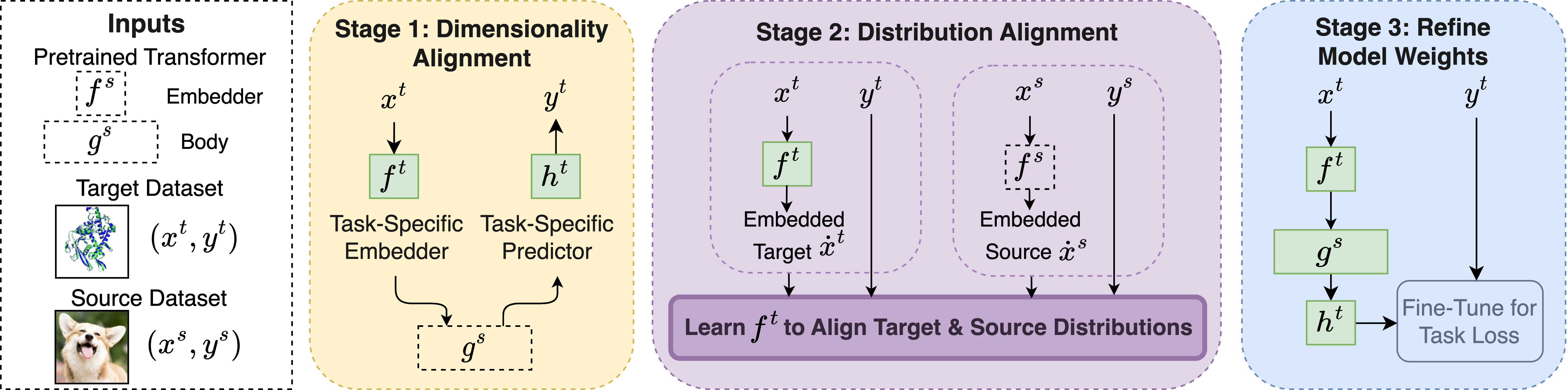}
 \vspace{-0.2cm}
  \caption{\small 
  \Algo's three-stage fine-tuning workflow  
  enables fast and automatic exploitation of large-scale pretrained models  for solving diverse tasks.  In stage 1, given target data $(x^t, y^t)$ and a pretrained transformer body $g^s$, \Algo constructs an
  embedder architecture $f^t$ to map the input to the dimensionality of $g^s$, and a predictor architecture $h^t$ to convert the output of  $g^s$  to the target output, e.g., classification logits. The weights of $f_t$ and $h_t$ are randomly initialized.
 In stage 2, 
\Algo learns  $f^t$ by minimizing the distributional distance between the  embedded target features and some in-modality source features. In stage 3, \Algo fine-tunes  $f^t$, $g^s$, and $h^t$ to minimize the task loss. 
}
 \vspace{-0.4cm}
\label{fig:workflow}
\end{figure*}

The rise of large-scale pretrained models has been a hallmark of machine learning (ML) research in the past few years. Using transfer learning, these models can apply what they have learned from large amounts of unlabeled data to downstream tasks and perform remarkably well in 
a number of modalities, such as  
language, vision, and speech processing \citeg{Radford2018ImprovingLU, Carion2020EndtoEndOD, Baevski2020wav2vec2A}.
Existing research focuses on \textit{in-modality transfer} within these well-studied areas---for example, BERT models \citep{Devlin2019BERTPO} are typically only adapted for text-based tasks, and vision transformers \citep{Dosovitskiy2021AnII} only for image datasets.

But imagine if we could use pretrained BERT models to tackle genomics tasks, or vision transformers to  solve PDEs? Effective \emph{cross-modal fine-tuning} could have immense impact on less-studied areas, such as  physical and life sciences, healthcare, and finance. Indeed, designing 
specialized  networks in  these areas is   challenging, as it requires both domain knowledge and ML expertise. Automated machine learning (AutoML)
\citeg{roberts2021rethinking,shen2022efficient}
and general-purpose architectures \citeg{jaegle2022perceiver} can be used to simplify this process, but they still require training models from scratch, which is  difficult for data-scarce modalities. 
Applying models pretrained in data-rich modalities to these new problems can potentially alleviate the modeling and data concerns, reducing the human effort needed to develop high-quality task-specific models.

Despite the potential  impact, the general feasibility of cross-modal fine-tuning remains an open question. While recent work has demonstrated its possibility by applying pretrained language models 
to vision tasks \citep{Dinh2022LIFTLF, Lu_Grover_Abbeel_Mordatch_2022}, referential games \citep{Li2020EmergentCP}, and reinforcement learning \citep{Reid2022CanWH}, many of these  approaches are ad-hoc, relying on manual prompt engineering or 
architecture add-ons to solve specific tasks. Besides, they often do not 
yield models that are competitive with those trained from scratch.  We aim to tackle both of these shortcomings.

In this work, we propose a fine-tuning workflow called \textbf{\Algo}
that bridges the gap between generality and effectiveness  in cross-modal learning.
Our key insight is to perform task-specific data alignment prior to task-agnostic fine-tuning. By matching the data distribution of an unfamiliar modality with that of a familiar one, \Algo can   prevent  the distortion of the pretrained weights and 
exploit the knowledge encoded in the pretrained models, achieving  significantly better results than naive fine-tuning and state-of-the-art performance on 3 benchmarks---NAS-Bench-360~\citep{nasbench360}, PDEBench~\citep{Takamoto2022PDEBENCHAE}, and OpenML-CC18~\citep{Vanschoren2014OpenMLNS}---which contain over 60 datasets from 12 distinct data modalities. 

Concretely, \Algo adapts any pretrained transformer model to a downstream task via a three-stage workflow (Figure~\ref{fig:workflow}). 
First, \Algo generates a task-specific embedding network architecture that maps the target  inputs  to sequence features which can be processed by the  transformer layers (\textit{dimensionality alignment}).
Then, the embedding network is trained to minimize the distributional distance between the embedded target features and the features of an in-modality reference dataset\footnote {Due to privacy and computational efficiency concerns, we do not assume access to the pretraining data and instead work with publicly available proxy data, e.g., CIFAR-10 for vision models.}  (\textit{distribution alignment}). Finally, the entire target model is fine-tuned to calibrate the weights with the task goal. In Section~\ref{sec:method:eval}, we evaluate several standard distance metrics for distribution alignment and find that the optimal transport dataset distance \citep{AlvarezMelis2020GeometricDD} attains the best empirical performance, possibly by taking the  label distribution and clustering structure of the data into consideration. Thus, we use it  in our subsequent experiments.

We validate \Algo's effectiveness along three axes: breadth, depth, and comparison with existing work. Breadthwise, we evaluate \Algo on  NAS-Bench-360 \citep{nasbench360}, an AutoML benchmark that includes 10  tasks with diverse input dimensions (1D and 2D), prediction types (point and dense), and modalities (vision, audio, electrocardiogram, physics, protein, genomics, and cosmic-ray). The   empirical results, combined with our analysis, show the following:
\vspace{-0.35cm}
\begin{itemize}[leftmargin=*,noitemsep]\setlength\itemsep{3pt}
    \item \textbf{Cross-modal fine-tuning is promising}: \Algo outperforms various  hand-designed  models,  AutoML methods, and general-purpose architectures, ranking first on 7 tasks and in the top three on all tasks. We also observe \Algo's effectiveness in  a simulated limited-data setting.
    \item \textbf{Alignment is crucial}: We find an empirical correlation between alignment quality and downstream accuracy. The fact that \Algo significantly outperforms naive fine-tuning demonstrates that   data alignment is important.
    \item \textbf{Alignment can be performed efficiently}: Our embedder learning time is only $\sim$10\% of the fine-tuning time.
\end{itemize}
\vspace{-0.3cm}
Depthwise, we study two established benchmarks in practical modalities: PDEBench for solving partial differential equations \citep{Takamoto2022PDEBENCHAE}  and OpenML-CC18  for classifying tabular  data \citep{Vanschoren2014OpenMLNS}. We perform in-depth analysis to show that \Algo adapts vision and language transformers to learn meaningful representations of the target tasks.
It matches the performance of state-of-the-art approaches, including FNO \citep{li2021fno} for PDEBench, AutoGluon \citep{Erickson2020AutoGluonTabularRA} and TabPFN \citep{Hollmann2022TabPFNAT} for OpenML-CC18.

Finally, we compare with  task-specific cross-modal methods that convert tabular data into text \citep{Dinh2022LIFTLF} or images \citep{Zhu2021ConvertingTD} to reuse existing models. The results clearly suggest that \Algo is both more effective and more general. Our code is made public at \url{https://github.com/sjunhongshen/ORCA}.

\section{Related Work}
\label{sec:related}

\begin{table*}[t]
\vspace{-0.2cm}
\caption{\small Summary of existing approaches for model development for diverse tasks.
}
\vspace{-0.1cm}
\label{table:relatedwork}
\centering
\vspace{-0.15cm}
\resizebox{0.85\textwidth}{!}{	\begin{tabular}{lcccccc}
\toprule
			 && Task-specific & General-purpose     &\multicolumn{3}{c}{Supports transfer to different: } \\
 	     && adaptation?   & workflow?           & input dim? & output dim? & modality? \\
\cmidrule(lr){1-4}\cmidrule(lr){5-7}
Task-specific &  Hand-designed models                       & \checkmark &            &     &  \\ 
    learning  & AutoML models& \checkmark & \checkmark &  \\  
\cmidrule(lr){1-4}\cmidrule(lr){5-7}
\multirow{3}{80pt}{In-modality transfer}
              & Unimodal DA                & \checkmark &            & \checkmark & \\
    & Uni/Multimodal fine-tuning    & \checkmark &            & \checkmark & \checkmark \\
              & General-purpose models     & \checkmark & \checkmark & \checkmark & \checkmark \\  
              
\cmidrule(lr){1-4}\cmidrule(lr){5-7}
\multirow{4}{80pt}{Cross-modal transfer}
              & Heterogeneous DA           & \checkmark &            & \checkmark &  & \checkmark\\
              & Task-specific fine-tuning  & \checkmark &            & \checkmark & \checkmark & \checkmark\\
    & FPT &            & \checkmark & \checkmark & \checkmark& \checkmark \\
		   & \textbf{\Algo }            & \checkmark & \checkmark & \checkmark & \checkmark& \checkmark \\
\bottomrule 
\end{tabular}}
\vspace{-0.25cm}
\end{table*}

In this section, we review several groups of related work in the areas of AutoML, in-modality transfer,
and cross-modal transfer.
Table \ref{table:relatedwork} summarizes these groups along relevant axes, and contrasts them with \Algo. 

\looseness=-1
\textbf{AutoML for diverse tasks} is a growing research area, as evidenced by the NAS-Bench-360 benchmark~\citep{nasbench360}, the \href{https://www.cs.cmu.edu/~automl-decathlon-22/}{2022 AutoML Decathlon competition}, and  recent neural architecture search (NAS) methods that target this problem, such as AutoML-Zero~\citep{Real2020AutoMLZeroEM}, XD~\citep{roberts2021rethinking}, and DASH~\citep{shen2022efficient}. Unlike   NAS methods which  repeatedly incur the overhead of designing new architectures and train them from scratch, \Algo takes a fine-tuning approach and reuses existing  models in data-rich modalities. That said, given the shared underlying motivation, we  use   NAS-Bench-360 in our experimental evaluation and  compare  against state-of-the-art AutoML baselines. 

\textbf{Unimodal domain adaptation (DA)} is a form of transductive transfer learning where \textit{the source and target tasks are the same} but the domains differ \citep{pan2009survey, wang2018deep}. Most DA methods assume that  the source and target data have the same input space and support, and are concerned with different output spaces or  joint/marginal distributions.
Recent work studies more general settings such as different feature spaces (heterogeneous DA) or label spaces (universal DA).  
Our focus on cross-modal fine-tuning goes one step further to the case where neither the input-space nor the output-space support overlaps.

\textbf{Unimodal fine-tuning} is a more flexible transfer approach that can be applied to downstream tasks with different label or input spaces.  Pretrained models 
are used for in-modality fine-tuning in fields like language \citeg{Jiang2020SMARTRA, Aghajanyan2021BetterFB}, vision \citeg{Li2022MaskDT, Wei2022ContrastiveLR}, speech \citeg{jiang2021further, chen2022wavlm},
protein \citep{Jumper2021HighlyAP}, and robotics \citep{Ahn2022DoAI}. 
Adapter networks \citep{He2022TowardsAU} have been developed to improve the performance of in-modality fine-tuning. 
\textbf{Multimodal fine-tuning} expands the applicable modalities of a single pretrained model by
learning embeddings of several modalities together \citeg{ Radford2021LearningTV, Hu2021UniTMM, Kim2021ViLTVT, Alayrac2022FlamingoAV}, but these methods still focus on adapting to \textit{in-modality} tasks.

\textbf{General-purpose models} propose flexible architectures applicable to various tasks such as 
optical flow, point clouds, and reinforcement learning \citep{Jaegle2021PerceiverGP, jaegle2022perceiver, Reed2022AGA}. These approaches train 
multitask transformers \textit{from scratch} using a large body of data from different tasks. Though more versatile than unimodal models, they still focus on \textit{transferring to problems within the considered pretraining modalities}. Nonetheless, the success of transformers for in-modality fine-tuning motivates us to focus on adapting transformer architectures for cross-modal tasks.

\looseness=-1
\textbf{Heterogeneous DA (HDA)}
considers nonequivalent feature spaces between the source and target domains. While most HDA methods tackle same-modality-different-dimension transfer, e.g., between images of different resolutions, there are indeed a few works studying cross-modal  text-to-image transfer \citep{yao2019heterogeneous, li2020simultaneous}. However, a crucial assumption that HDA makes is that the \textit{target and source tasks are the same}. In contrast, we consider more flexible knowledge transfer between drastically different modalities with \textit{distinct tasks and label sets}, such as applying Swin Transformers to solving partial differential equations or RoBERTa to classifying electrocardiograms.

\textbf{Cross-modal task-specific fine-tuning} is a recent line of research, with most work focusing on
transferring  language models to other modalities like vision \citep{Kiela2019SupervisedMB}, referential games \citep{Li2020EmergentCP},  reinforcement learning \citep{Reid2022CanWH}, and protein sequences \citep{Vinod2023ReprogrammingPL}. These works provide initial evidence of the cross-modal transfer capacity of pretrained models.
However, they focus on hand-tailoring to a single modality, e.g., by adding ad-hoc encoders that transform agent messages \citep{Li2020EmergentCP} or decision trajectories \citep{Reid2022CanWH} into tokens. 
Even when not relying on fine-tuning, work like LIFT \citep{Dinh2022LIFTLF} that attempts cross-modal learning via  prompting \citep{liu2021pre} still requires ad-hoc conversion of tasks  to natural text.

\textbf{Frozen Pretrained Transformers (FPT)} \citep{Lu_Grover_Abbeel_Mordatch_2022} is a  cross-modal fine-tuning workflow that transforms the inputs to be compatible with the pretrained models. Although FPT and \Algo are both general-purpose, FPT does not account for the modality difference  (no stage 2 in Figure~\ref{fig:workflow}), but we show this step is necessary to obtain effective predictive models and outperform existing baselines.

\section{\Algo Workflow}

In this section, we  formalize the problem setup and  introduce  the our workflow for adapting pretrained transformers.

\textbf{Problem Setup.}
A domain $\mathcal{D}$ consists of a feature space $\mathcal{X}$, a label space $\mathcal{Y}$, and a joint probability distribution $P(\mathcal{X}, \mathcal{Y})$.  
In the cross-modal setting we study, the target (end-task) domain $\mathcal{D}^t$ and source (pretraining) domain $\mathcal{D}^s$ differ not only in the feature space but also the label space and by extension have differing probability distributions, i.e., $\mathcal{X}^t\neq\mathcal{X}^s$, $\mathcal{Y}^t\neq\mathcal{Y}^s$, and  
$P^t(\mathcal{X}^t, \mathcal{Y}^t)\neq P^s(\mathcal{X}^s, \mathcal{Y}^s)$. This is in contrast to the  transductive transfer learning setting addressed by domain adaptation, where source and target domains share the label space and end task \citep{pan2009survey}.

\looseness=-1
Given target data $\{x_i^t, y_i^t\}_{i=1}^{n^t}$ sampled from a joint distribution $P^t$ in domain $\mathcal{D}^t$, our goal is to learn a  model $m^t$ that correctly maps each input $x^t$ to its label $y^t$.
We are interested in achieving this using pretrained transformers. Thus, we assume access to a model $m^{s}$ pretrained  with data  $\{x_i^s, y_i^s\}_{i=1}^{n^s}$ in the source domain $\mathcal{D}^s$.
Then, given a  loss function $l$, we aim to develop $m^t$ based on $m^s$ such that $\mathbb{E}_{(x^t, y^t)\sim P^t} [l(m^t(x^t), y^t)]$ is minimized. This problem formulation does not define modality explicitly and includes both in-modal and cross-modal transfer.  Given the generality of the tasks we wish to explore and the difficulty of differentiating the two settings mathematically,  we rely on semantics to do so: intuitively, cross-modal data (e.g., natural images vs. PDEs) are more distinct to each other than in-modal data (e.g., photos taken in two  geographical locations).

Having defined the learning problem, we now present our three-stage cross-modal fine-tuning workflow: (1) generating task-specific embedder and predictor to support diverse input-output dimensions, (2) pretraining embedder  to align the source and target feature distributions, and (3) fine-tuning to minimize the target  loss.

\subsection{Architecture Design for Dimensionality Alignment}
\label{sec:architecturedesign}

Applying pretrained models to a new problem usually requires addressing the problem of dimensionality mismatch.
To make \Algo work for   different input/output dimensions, we decompose a transformer-based learner $m$ into three parts (Figure~\ref{fig:workflow} stage 1): an embedder $f$ that transforms input $x$ into a sequence of features, a model body $g$ that applies a  series of pretrained attention layers to the embedded features, and a predictor $h$ that generates the outputs with the desired  shape. \Algo uses a pretrained architecture and weights to initialize the model body $g$ but replaces $f$ and $h$ with layers designed to match the target data with the pretrained model's embedding dimension. 
In the following, we describe each module in detail.

\textbf{Custom Embedding Network.}
Denote the feature space compatible with the pretrained model  as $\mathcal{\dot X}$. For a  transformer with maximum sequence length $S$ and embedding dimension $D$, $\mathcal{\dot X}=\mathbb{R}^{S\times D}$.
The target embedder $f^t: \mathcal{X}\rightarrow \mathcal{\dot X}$ is designed to take in a tensor of arbitrary dimension from $\mathcal{X}$ and transform it to $\mathcal{\dot X}$.
In \Algo, $f^t$ is composed of a convolutional layer with input channel $c_{in}$, output channel $c_{out}$, kernel size $k$, and stride $k$, generalizing the patching operations used in vision transformers to 1D and higher-dimensional cases. We set $c_{in}$ to the input channel of  $x$ and $c_{out}$ to the embedding dimension $D$. 
We can either treat $k$ as a  hyperparameter or set it to the smallest value for which the product of output shape excluding the channel dimension $\leq S$ to take full advantage of the representation power of the pretrained model. In the latter case, when we flatten the non-channel dimensions of the output tensors after the convolution, pad and then transpose it, we can obtain sequence features with shape $S\times D$.
Finally, we add a layer norm and a positional embedding to obtain  $\dot x$.

\textbf{Pretrained Transformer Body.}
The model body $g$ takes the embedding $\dot x\in \mathcal{\dot X}$ as input and outputs features $\dot y\in \mathcal{\dot Y}$; the dot is used to differentiate these intermediate representations from the raw inputs and labels.
For transformer-based $g$, both the input and output feature spaces $\mathcal{\dot X}, \mathcal{\dot Y}$ are $\mathbb{R}^{S\times D}$. 

\textbf{Custom Prediction Head.}
Finally, the target model's prediction head $h^t$ must take $\dot y\in \mathcal{\dot Y}$ as input and return a task-dependent output tensor.
Different tasks often specify different types of outputs, e.g., classification logits in $\mathbb{R}^K$, where $K$ is the number of classes, or dense maps where the spatial dimension is the same as the input and per index logits correspond to $K$ classes. 
Thus, it is crucial to define task-specific output modules and fine-tune them for new problems. In \Algo, we use the simplest  instantiation of the predictors. For classification, we apply average pooling along the sequence length dimension
to obtain 1D tensors with length $D$ and  then use a linear layer that maps $D$ to $K$. For dense prediction, we apply a linear layer to the sequence outputs so the resulting tensor has shape $(S, k^{\mathrm{ndim}(\mathcal{Y})}K)$, where $k^{\mathrm{ndim}(\mathcal{Y})}$ is the downsampling factor of the embedder convolution kernel with stride $k$. This upsamples by the same factor that the embedder  downsampled.  Then, we can mold the tensor to the desired output dimension\footnote{For example, consider an image   with shape $(C_{in}, H_{in}, W_{in})$.  We  choose   $k$ for the embedder such that $H_{out}\times W_{out} \approx S$ so the output  shape is $(D, H_{out}, W_{out})$. Then, we flatten the last two dimensions and transpose to get shape $(S, D)$  compatible with the transformer. The transformer output   is mapped to $(S, k^2K)$ by a linear layer. We  transpose  and reshape to get $(k^2K, H_{out}, W_{out})$ and apply pixelshuffle~\citep{shi2016real}  to get   $(K, H_{in}, W_{in})$.}.

With an architecture based on the pretrained model but also compatible with the target task, we can now turn our attention to data alignment for better adaptation.

\renewcommand{\res}[2]{#1}
\begin{table*}[t]
\vspace{-0.2cm}
  	\caption{\small Prediction errors ($\downarrow$) on  10 diverse tasks. 	``NAS-Bench-360" refers to the task-wise best of all AutoML baselines evaluated in the paper, including DARTS~\citep{liu2018darts}, DenseNAS~\citep{Fang2020DenselyCS},  and 4 others. ``FPT" refers to fine-tuning the layer norms of RoBERTa/Swin. \textbf{On 7/10 problems, \Algo ranks the first among all competitors. }
		 See Appendix~\ref{appendix:nb360std} for the error bars.
		}
		\large
		\vspace{-0.2cm}
		\label{table:accwitherror}
		\large
\resizebox{1.0\textwidth}{!}{	\begin{tabular}{lcccccccccc}
			\toprule 
			 & CIFAR-100 & Spherical  & Darcy Flow & PSICOV & Cosmic& NinaPro & FSD50K  & ECG & Satellite & DeepSEA  \\
			  & 0-1 error (\%)  & 0-1 error (\%) & relative $\ell_2$ & MAE$_8$  & 1-AUROC   & 0-1 error (\%)   & 1- mAP & 1 - F1 score  & 0-1 error (\%)  & 1- AUROC \\
			
			\midrule 
			Hand-designed  &\res{19.39}{0.20}  & \res{67.41}{0.76}  &\res{8E-3}{1E-3} & \res{3.35}{0.14} &  \textbf{\res{0.127}{0.01}} &\res{8.73}{0.90}  & \res{0.62}{0.004} & \textbf{\res{0.28}{0.00} } & \res{19.80}{0.00} & \res{0.30}{0.024} \\ 

            \midrule
		  NAS-Bench-360&\res{23.39}{0.01}  & \res{48.23}{2.87} & \res{2.6E-2}{1E-3}  & \res{2.94}{0.13} & \res{0.229}{0.04} & \res{7.34}{0.76} & \res{0.60}{0.001} &\res{0.34}{0.01}& \res{12.51}{0.24} &  \res{0.32}{0.010} \\ 
			DASH &\res{24.37}{0.81}  & \res{71.28}{0.68}  & \res{7.9E-3}{2E-3}  & \res{3.30}{0.16} &\res{0.19}{0.02} &\textbf{\res{6.60}{0.33}} & \res{0.60}{0.008}  & \res{0.32}{0.007}  & \res{12.28}{0.5} & \textbf{\res{0.28}{0.013}}\\
			\midrule 	
			Perceiver IO  & \res{70.04}{0.44} & \res{82.57}{0.19} & \res{2.4E-2}{1E-2} & \res{8.06}{0.06} & \res{0.485}{0.01} & \res{22.22}{1.80} & \res{0.72}{0.002} &\res{0.66}{0.01}& \res{15.93}{0.08} & \res{0.38}{0.004} \\
			FPT &\res{10.11}{1.18} & \res{76.38}{4.89}&\res{2.1E-2}{1.3E-3} & \res{4.66}{0.054} &\res{0.233}{0.002}& \res{15.69}{2.33}&\res{0.67}{0.0068} &\res{0.50}{0.0098} & \res{20.83}{0.24}&\res{0.37}{0.0002} \\
			\midrule
		\textbf{\Algo}  &\textbf{\res{6.53}{0.079}}  & \textbf{\res{29.85}{0.72}}&\textbf{\res{7.28E-3}{6.8E-5}} & \textbf{\res{1.91}{0.038}} &\res{0.152}{0.005}& \res{7.54}{0.39}&\textbf{\res{0.56}{0.013}} &\textbf{\res{0.28}{0.0059}}&\textbf{ \res{11.59}{0.18}}  & \res{0.29}{0.006}  \\
			\bottomrule 
		\end{tabular}}
		\vspace{-0.1cm}
	\end{table*}
\renewcommand{\res}[2]{#1$\pm$#2}

\subsection{Embedder Learning for Distribution Alignment}
\label{sec:method:otdd}

\looseness=-1
Intuitively, transferring knowledge across similar modalities should be easier than across distant ones. Hence, given a target task in a new modality, we aim to manipulate 
the target data so that they become closer to the pretraining modality. 
One way to achieve this is to train the embedder before actually fine-tuning the model body in a way that makes the \textit{embedded target features} resemble the source features which the pretrained model body is known to perform well on.

Formally, let $f^s: \mathcal{X}^s \to \mathcal{\dot X}$ denote the pretrained source embedder (the part of $m^s$ that transforms the raw data to sequence features) and $f^t$ the randomly initialized target embedder 
discussed in the previous section.  
We can learn $f^t$ to minimize the  distance between the joint  distribution of the target embeddings $\big(f^t(x^t), y^t\big)$ and that of the source embeddings  $\big(f^s(x^s), y^s\big)$. 
There are many metrics for measuring this distributional distance.
To understand whether they affect adaptation differently, 
we perform a preliminary study in Section~\ref{sec:method:eval} on three representatives.

\subsection{Weight Refining for Downstream Adaptation}
After training the embedder, we perform full fine-tuning by updating all model parameters to minimize the target  loss. This step further aligns the embedder and predictor with the pretrained model.
In Section~\ref{sec:exp:finetuneablation}, we compare \Algo with standard fine-tuning without data alignment  and show that our approach  improves performance while reducing  variance. There are orthogonal  works that study how to best fine-tune a  model \citeg{Liu2022PretrainPA, He2022TowardsAU}. We compare with one strategy used in FPT \citep{Lu_Grover_Abbeel_Mordatch_2022} in Section~\ref{sec:exp:layernormablation} but leave further exploration for future work.

\subsection{Evaluation of Distribution Alignment Metrics}
\vspace{-0.1cm}
\label{sec:method:eval} 
We evaluate the effectiveness of three distance metrics for data alignment during embedding learning: (1) the pairwise Euclidean distance, which  aligns the scales and ranges of the datasets without using any distributional information; (2) the moment-based maximum mean discrepancy (MMD) \citep{Gretton2012AKT}, which uses the  distribution of $f(x)$ to align the feature means;  and  (3) optimal transport dataset distance (OTDD) \citep{AlvarezMelis2020GeometricDD}, which uses both the feature and  label distributions $\big(f(x), y\big)$ to align the high-level clustering structure of the datasets. 
  
\begin{figure}[t]
\vspace{-0.4cm}
\centering
    \includegraphics[width=0.25\textwidth]{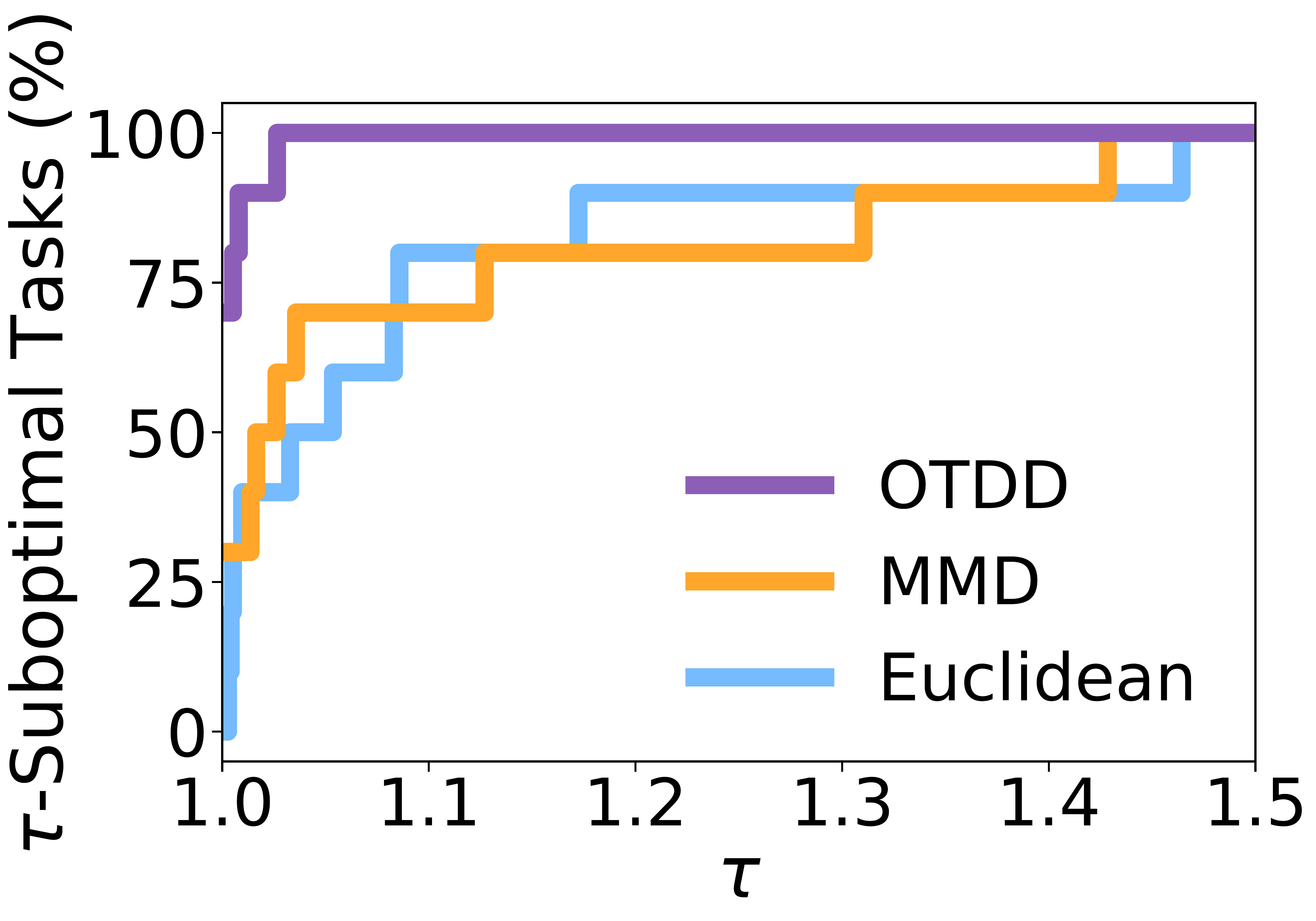}
    \vspace{-0.4cm}
    \caption{\small Performance profiles \citep{dolan2002profiles} of 
    \Algo with different alignment metrics. Larger values (fractions of tasks on which a method is within $\tau$-factor of the best) are better. {The OTDD curve  being in the upper left  shows it is often the best.} 
    }\label{fig:profile_metric}
    \vspace{-0.32cm}
  \end{figure}

We  substitute each metric into the \Algo workflow (implementation details  in Section~\ref{sec:experiments}) and evaluate them on 10 tasks from diverse modalities (benchmark details in Section~\ref{sec:exp:layernormablation}). {The aggregate performance (Figure~\ref{fig:profile_metric}) and  per-task rankings (Appendix~\ref{appendix:metrics}) show that  embedder learning with 
OTDD has the best overall results, so we use it in our subsequent experiments. We conjecture that its good performance is due to how the label information is considered during alignment.}

Indeed, for both the source and target datasets, OTDD represents each class label as a distribution over the in-class features: $y \mapsto  P(\mathcal{\dot X} | \mathcal{Y} = y)$\footnote{This step requires that the labels be discrete, as in the classification datasets. For dense prediction tasks with continuous labels, we first perform clustering on the data labels to generate pseudo-labels.}. This transforms the source and target label sets into the shared space of distributions over $\mathcal{\dot X}$. Then, we can define the distance $d_\mathcal{Y}(y^t, y^s)$ between different labels using the $p$-Wasserstein distance associated with the $l_2$ 
distance $\|\dot x^t -\dot x^s\|_2^2$ in $\mathcal{\dot X}$, which in turn allows us to measure the distributional difference in $\mathcal{\dot X} \times \mathcal{Y}$:
\vspace{-0.2cm}
\begin{align*}
    d_{\mathcal{\dot X} \times \mathcal{Y}}\big((\dot x^t, y^t), (\dot x^s, y^s)\big) = \big(d_{\mathcal{\dot X}}(\dot x^t, \dot x^s)^p + d_\mathcal{Y}(y^t, y^s)^p\big)^{1/p}.
\end{align*}
We refer the readers to \citet{AlvarezMelis2020GeometricDD} for the exact formulation.
Yet the implication from our experiments is that, as we learn $f^t$ to minimize OTDD, we are not only aligning individual data points, but also  grouping   features with the same label together in the embedding space, which could potentially facilitate fine-tuning. 

Despite its effectiveness for data alignment,  OTDD is generally expensive to compute. In Section~\ref{appendix:otdd} of the Appendix, we analyze its computational complexity   and propose an efficient approximation to it using class-wise subsampling. 

\looseness=-1
Before ending this section, we  emphasize that our goal is not to discover the best alignment metric but to\textit{ provide a general  fine-tuning framework that works regardless of the metric used}. Thus, we leave designing more suitable distance metrics for future work.

\section{Experiments}
\label{sec:experiments}

\renewcommand{\res}[2]{#1}
\begin{table*}[t]
 \vspace{-0.2cm}
  		\caption{\small Prediction errors ($\downarrow$) of \Algo, naive fine-tuning, and training RoBERTa/Swin from scratch. We consider adapting all parameters (full setting) vs. only the layer norms (FPT setting).\textbf{\Algo is better in both settings.} The fact that full fine-tuning  generally outperforms  tuning only the layer norms is also consistent with recent observations  
\cite{Rothermel2021DontSY}. See Appendix~\ref{appendix:layernormstd} for the error bars.
		}
		\large
		\vspace{-0.2cm}
		\label{table:layernorm}
\resizebox{1.0\textwidth}{!}{		\begin{tabular}{lcccccccccc}
			\toprule 
			 & CIFAR-100 & Spherical  & Darcy Flow & PSICOV & Cosmic& NinaPro & FSD50K & ECG & Satellite & DeepSEA  \\	
			\midrule 
			Train-from-scratch &\res{50.87}{0.32} & \res{76.67}{0.21}&\res{8.0E-2}{1.3E-2} &
			\res{5.09}{0.014}&\res{0.50}{0.00}&
			\res{9.96}{1.67}&\res{0.75}{0.017}&\res{0.42}{0.011} & \res{12.38}{0.14} & \res{0.39}{0.01}\\
				\midrule
				Fine-tuning &\res{7.67}{0.55} & \res{55.26}{1.63}&\res{7.34E-3}{1.1E-4} & \res{1.92}{0.039}&\res{0.17}{0.011}& \res{8.35}{0.75}&\res{0.63}{0.014} &\res{0.44}{0.0056} & \res{13.86}{1.47}&\res{0.51}{0.0001} \\
		\textbf{\Algo} &\textbf{\res{6.53}{0.079}}  & \textbf{\res{29.85}{0.72}}&\textbf{\res{7.28E-3}{6.8E-5}} & \textbf{\res{1.91}{0.038}} &\textbf{\res{0.152}{0.005}}&\textbf{ \res{7.54}{0.39}}&\textbf{\res{0.56}{0.013}} &\textbf{\res{0.28}{0.0059}}&\textbf{ \res{11.59}{0.18}}  & \textbf{\res{0.29}{0.006}}\\
				\midrule
			Fine-tuning (layernorm) &\res{10.11}{1.18} & \res{76.38}{4.89}&\res{2.11E-2}{1.3E-3} & \res{4.66}{0.054} &\res{0.233}{0.002}& \res{15.69}{2.33}&\res{0.67}{0.0068} &\res{0.50}{0.0098} & \res{20.83}{0.24}&\res{0.37}{0.0002}\\
		\textbf{\Algo (layernorm)} &\res{7.99}{0.098} & \res{42.45}{0.21}&\res{2.21E-2}{7.4E-4} & \res{4.97}{0.14} &\res{0.227}{0.003}& \res{15.99}{1.92}&\res{0.64}{0.0093}&\res{0.47}{0.007} & \res{20.54}{0.49} & \res{0.36}{0.0070}\\
			
			\bottomrule 
		\end{tabular}}
		\vspace{-0.3cm}
	\end{table*}
 \renewcommand{\res}[2]{#1$\pm$#2}

\looseness=-1
Having introduced how \Algo tackles cross-modal fine-tuning, we proceed with showing its
empirical efficacy via three thematic groups of experiments: (1) we evaluate \Algo across a \textit{breadth} of modalities and show that it 
outperforms  hand-designed, AutoML-searched, and general-purpose architectures; we   study its key components to  understand the mechanism behind cross-modal fine-tuning and exemplify how it benefits limited-data modalities; (2) we  perform  \textit{in-depth} analyses in two  modalities, PDE solving and tabular classification, to show that \Algo is competitive with expert-designed task-specific models;
(3) we compare \Algo  with previous ad-hoc  cross-modal learning techniques to show that we strike a balance between generality and effectiveness.

\textbf{Experiment Protocol. }
While our workflow accepts a wide range of pretrained transformers as model bodies, we use RoBERTa \citep{Liu2019RoBERTaAR} and Swin Transformers \citep{Liu2021SwinTH}, which are representatives of the most studied language and vision modalities, to exemplify \Algo's efficacy. We implement the  base models 
using the Hugging Face  library \citep{Wolf2019HuggingFacesTS} and choose  CoNLL-2003  and CIFAR-10 as the proxy datasets, respectively.
For each  task, we first perform hyperparameter tuning in the standard fine-tuning setting
to identify the optimal target sequence length, batch size, and optimizer configuration. 
Experiments are performed on a single NVIDIA V100 GPU and managed using the Determined AI platform. Results are  averaged over 5 trails. 
For other details, see Appendix~\ref{appendix:implementation}.

\subsection{A Breadth Perspective: Can Pretrained Models Transfer Across Modalities?}

\looseness=-1
In this section, we highlight the most important observation of this work: \textbf{cross-modal fine-tuning with data alignment can solve diverse tasks effectively and efficiently}. To show this, we test \Algo on 10 tasks from NAS-Bench-360\footnote{NAS-Bench-360 is designed for testing how well ML algorithms generalize and is a core component of the \href{https://www.cs.cmu.edu/~automl-decathlon-22/}{2022 AutoML Decathlon competition}. See Appendix~\ref{appendix:taskinfo} for the  task summary.} covering  diverse 1D/2D problems such as  protein folding,  cardiac disease prediction, and cosmic-ray detection. 
Following Table~\ref{table:relatedwork}, we consider 3 classes of baselines: (1) hand-designed, task-specific models identified by \citet{nasbench360};
(2)  general-purpose models represented by Perceiver IO~\citep{jaegle2022perceiver};  (3) AutoML methods, including the leading algorithm on NAS-Bench-360,  DASH \citep{shen2022efficient}.

 \begin{figure}[t]
\vspace{-0.2cm}
\centering
    \includegraphics[width=0.25\textwidth]{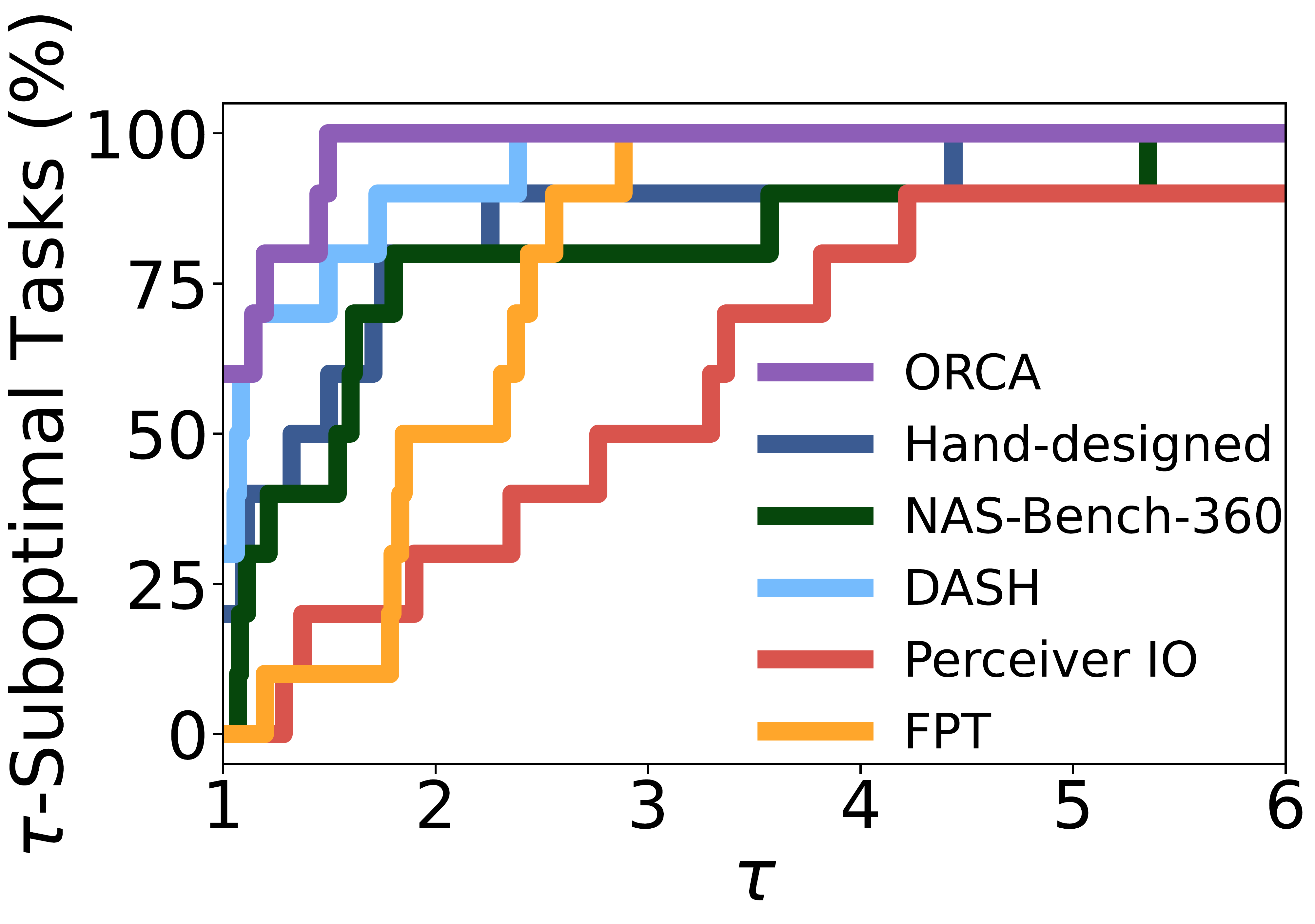}
    \vspace{-0.4cm}
    \caption{\small Aggregating Table~\ref{table:accwitherror} 
 results using performance profiles~\citep{dolan2002profiles}. Larger values (fractions of
tasks on which a method is  within $\tau$-factor
of the best) are better. \textbf{\Algo being in the
top left corner means it is often the best.} 
    }\label{fig:profile}
    \vspace{-0.6cm}
  \end{figure}
 \begin{figure*}[t]
   \begin{minipage}[c]{0.7\textwidth}
  \vspace{-0.1cm}
  \centering
  \includegraphics[width=\textwidth]{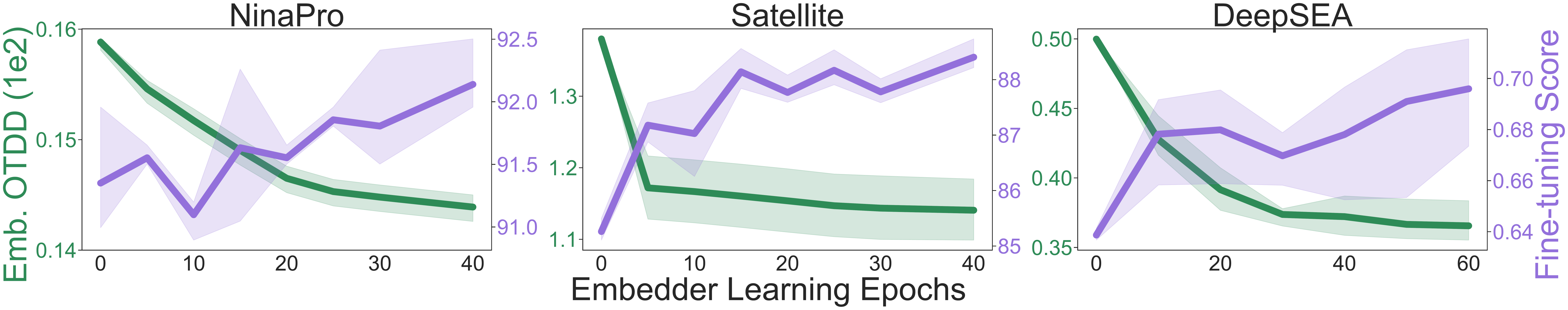}%
    \vspace{-3mm}
  \end{minipage}\hfill
\begin{minipage}[c]{0.28\textwidth}
\centering
    \includegraphics[width=0.75\textwidth]{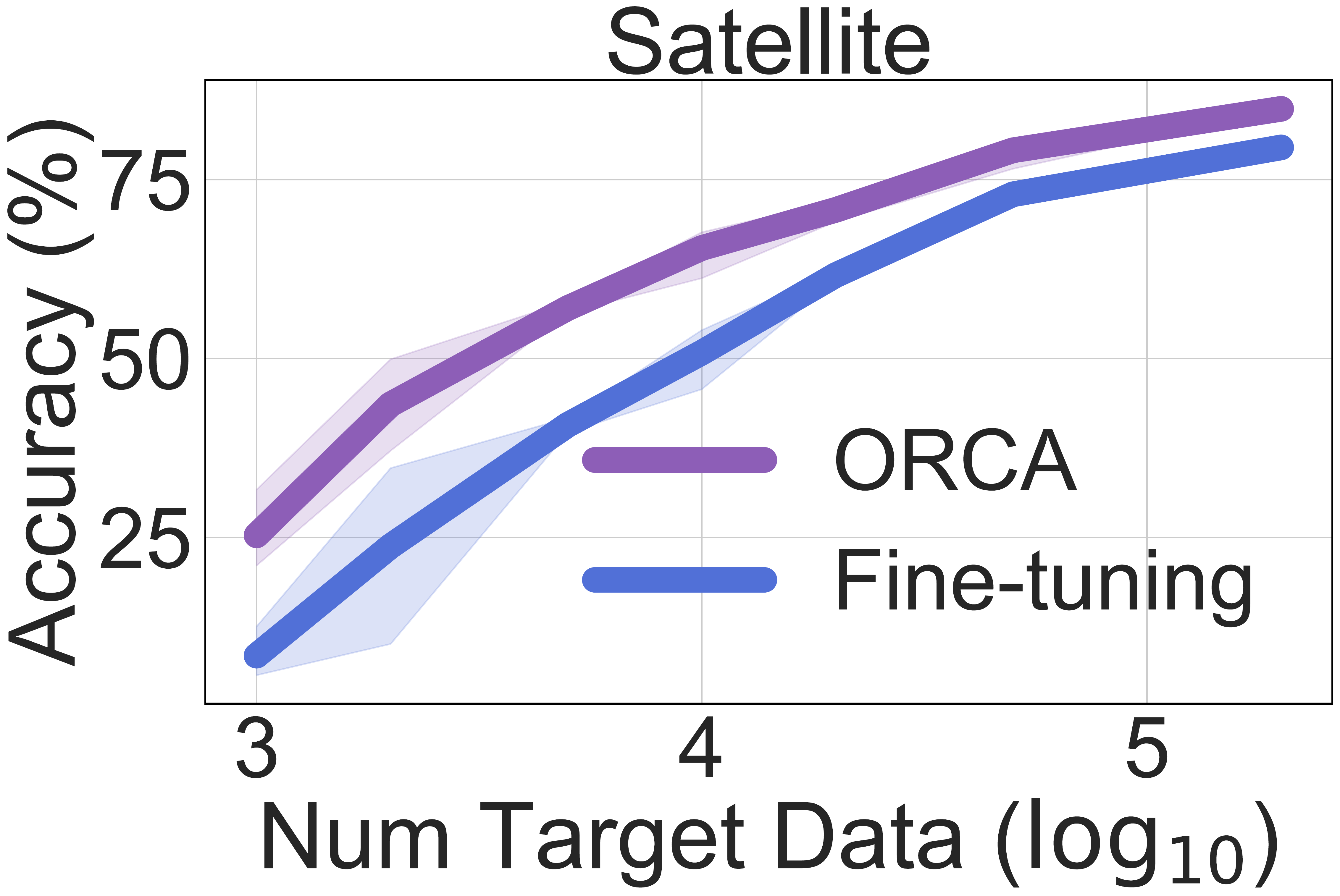}
    \vspace{-0.3cm}
  \end{minipage}
  \caption{\small \textbf{Left: }Final accuracy and embedding distribution distance vs. embedder learning epochs  on three NAS-Bench-360 tasks. As  we learn to map the target data to the source modality better (smaller OTDD), we  obtain  models with better downstream performance. This shows an empirical correlation between fine-tuning accuracy and alignment quality. \textbf{Right: }Accuracy ($\uparrow$) of \Algo vs. naive fine-tuning   with varying dataset size on task Satellite. \textbf{\Algo has  higher performance gains in low-data regime.}}
\label{fig:acc_dist}
\vspace{-0.2cm}
  \end{figure*}
  
We report the prediction error for each method on each task in Table \ref{table:accwitherror} and visualize  the aggregate performance in Figure~\ref{fig:profile}. \textbf{\Algo achieves the lowest error rates on 7 of 10 tasks and  the best aggregate performance}. Specifically, it outperforms hand-designed architectures on all tasks. It   beats all AutoML baselines on all tasks except DeepSEA and NinaPro, where it ranks second and third, respectively.
The improvements from the embedder learning stage of \Algo come at a small computational overhead---Table~\ref{table:timebreakdown} in the Appendix shows that the time needed for data alignment is only a small portion (11\%) of the fine-tuning time. 

Our results   validate the finding in prior cross-modal work that pretrained transformers  learn knowledge transferable to seemingly unrelated tasks. In the following, we dissect the success of \Algo via multiple ablations and identify 3 factors crucial to exploiting the learned knowledge: data alignment, full fine-tuning,  pretraining  modality selection.

\subsubsection*{Key 1: Aligning Feature Distributions}
\label{sec:exp:finetuneablation}

To understand whether the good performance of \Algo is indeed attributed to the data alignment process, which is our key innovation, we  compare it   with naive fine-tuning that does not align the data (Table~\ref{table:layernorm}, middle rows). We  see that \textbf{\Algo consistently outperforms naive fine-tuning}. {Moreover, we show in Appendix~\ref{appendix:metrics} that \Algo with different alignment metrics  all obtain better performance than   fine-tuning.} Thus, closing the gap between the target and pretraining modalities can facilitate model adaptation. 

To further isolate the impact of data alignment, we  compare \Algo with a train-from-scratch baseline (Table~\ref{table:layernorm}, first row) which trains  RoBERTa and Swin using only the target data.
We observe \textbf{training from scratch is worse than \Algo but better than fine-tuning} on ECG, Satellite, and DeepSea. We conjecture that this is because when the target modality differs significantly from the pretraining modality, naive fine-tuning may harm transfer, but aligning the feature distribution using \Algo can resolve this issue and benefit transfer. 
Indeed, recent work has shown that optimizing directly for the task loss may distort the pretrained weights and lead to suboptimal solutions \citep{Kumar2022FineTuningCD, Lee2022SurgicalFI}. By manipulating the target  distribution to look like the source distribution, we lower the risk of weight distortion, thus obtaining better downstream performance.

We also quantify the effect of data alignment by training the embedder for different number of epochs and see whether optimizing distribution distance to various levels of convergence affects downstream performance. Figure~\ref{fig:acc_dist} (left) plots the fine-tuning accuracy and the final distribution distance for different  embedder learning levels. We see that \textbf{as the dataset distance decreases, the fine-tuning accuracy increases}. 
In addition, learning the embedder separately from fine-tuning stabilizes training, as the performance variance of \Algo is constantly lower than that of naive fine-tuning. 
These results confirm that data alignment is the key to effective cross-modal fine-tuning.

\subsubsection*{Key 2: Fine-Tuning All Model Parameters}
\label{sec:exp:layernormablation}

\looseness=-2
As discussed in Section~\ref{sec:related}, Frozen Pretrained Transformers (FPT) \citep{Lu_Grover_Abbeel_Mordatch_2022} is a related work that showed pretrained language models contain knowledge relevant to out-of-modality tasks. While FPT presented a general pipeline for adapting GPT-2 to tasks like CIFAR-10,
the resulting models were not as good as those  trained from scratch. 
FPT differs from \Algo in that (1) it does not perform data alignment, and (2) it only fine-tunes the layer norms.  
We have verified the importance of (1). Now, we isolate the impact of (2) by fine-tuning only the layer norms for \Algo.

The bottom rows of Table \ref{table:layernorm} show that \Algo with fine-tuning  the layer norms outperforms FPT, so pretraining the embedder can boost the  performance of FPT.  However, this performance gain is smaller than that  in the full fine-tuning setting, which implies that \textbf{full fine-tuning can take better advantage of the learned embeddings}.  
In terms of runtime, FPT yields less than a 2$\times$ speedup compared with full fine-tuning (Appendix~\ref{appendix:runtimefpt}), despite the fact that we are updating many fewer parameters. This is unsurprising since gradients are still back-propagated through the entire network. Therefore, when computation allows, we recommend using \Algo with full fine-tuning for better  performance.

\subsubsection*{Key 3: Adapting From the Right  Modality}

Finally, we study how the pretraining modality affects fine-tuning. In the results reported so far,
we choose pretrained models for each task based on the input dimension, i.e., we use RoBERTa for all 1D tasks and Swin for all 2D tasks. Now, we  evaluate the opposite approach, focusing on two tasks: DeepSEA (1D) and Spherical (2D). This evaluation is straightforward to perform  by switching the model bodies, since
the embedder architecture of \Algo handles all input transformations needed to obtain the sequence features. The results are shown in Table~\ref{table:modality} in the Appendix. We see that fine-tuned RoBERTa outperforms Swin on the 1D task, possibly because the  DeepSEA data (genomics sequences) are structured more like language than images with discrete units of information and general grammatical rules. More crucially, 
for both  tasks, \textbf{models with smaller final OTDDs have better fine-tuning accuracy}. 
This suggests a way of selecting pretrained models 
by comparing the optimized OTDDs and picking the one with the smallest value.

\looseness=-1
Apart from these three key insights, recall that one of our  motivations for cross-modal fine-tuning is to help tasks with limited data, where training models from scratch is difficult. 
Indeed, for vanilla fine-tuning, a small amount of data may not give enough signal to update the pretrained weights, but it is possible to  learn a good embedder first with \Algo, which can then make fine-tuning easier. 
In Figure~\ref{fig:acc_dist} (right), we vary the dataset size and find that \textbf{the performance gain of \Algo increases as the dataset size  decreases}. 
Meanwhile, using \Algo allows us to match the performance of naive fine-tuning on 3$\times$ amount of data. 
Thus, it can benefit model development in domains where data collection is costly. Beyond the cross-modal setting, we  also verify \Algo's efficacy for in-modality transfer in Appendix~\ref{appendix:inmodal}.

\vspace{0.2cm}
\subsection{A Depth Perspective: Cross-Modal Fine-Tuning for PDE  and Tabular Tasks}

\begin{figure*}[t]
   \begin{minipage}[c]{0.67\textwidth}
  \centering
   \includegraphics[width=\textwidth]{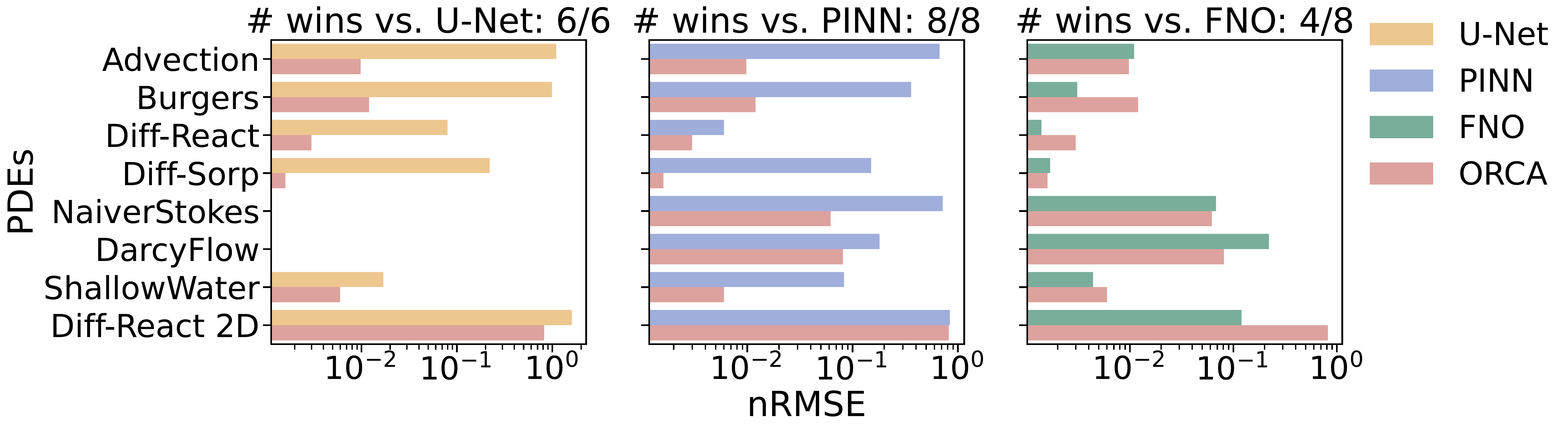}%
    \vspace{-2mm}

  \end{minipage}\hfill
\begin{minipage}[c]{0.3\textwidth}
\vspace{-0.2cm}
\centering
    \includegraphics[width=0.9\textwidth]{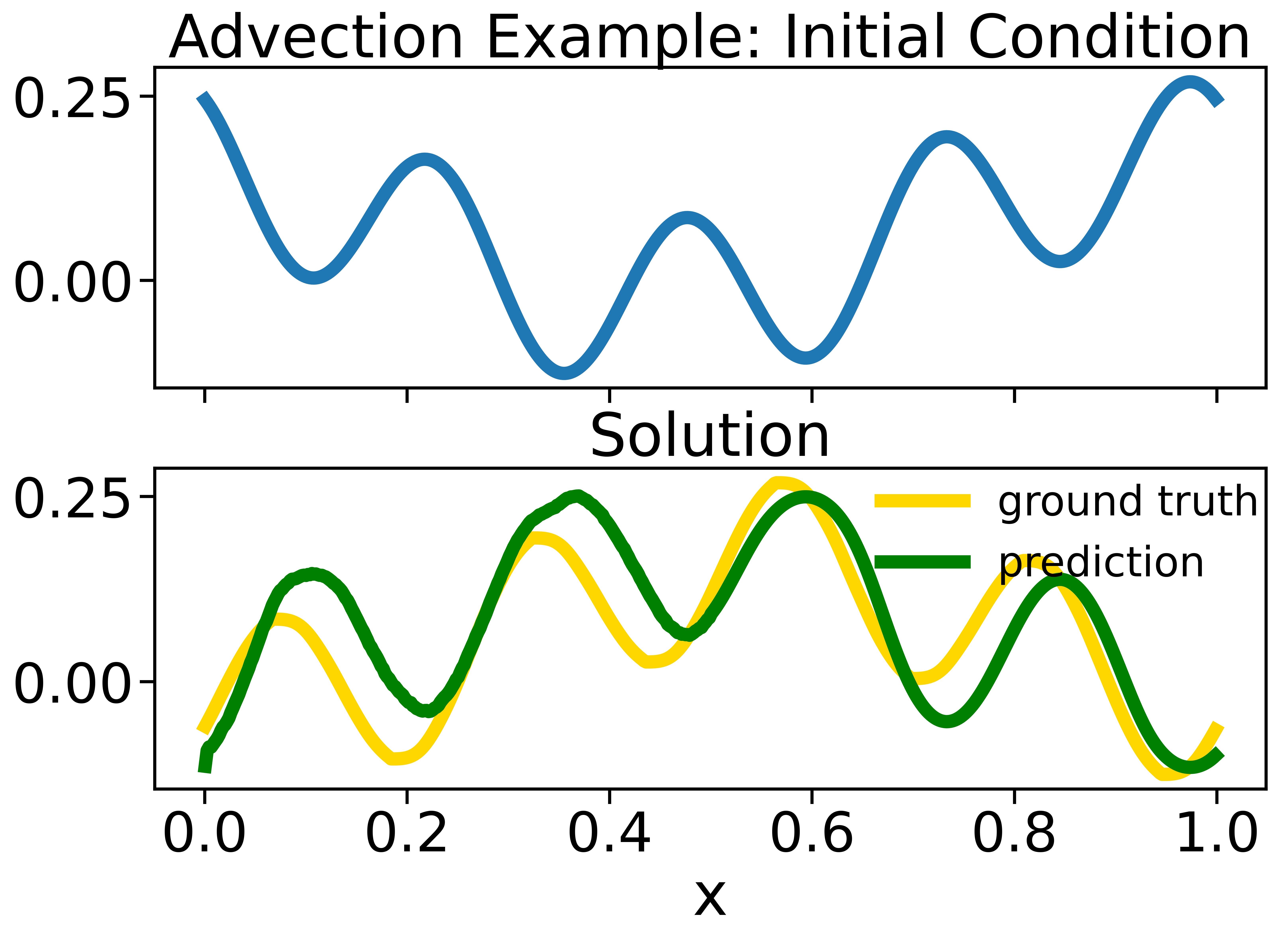}
    \vspace{-0.35cm}
  \end{minipage}
  \vspace{-0.2cm}
      \caption{\small  \textbf{Left: }Normalized Root Mean Squared Errors (nRMSEs, $\downarrow$) for \Algo vs. baselines on 8 PDEBench tasks with varying dimensions (1D/2D). We only evaluate datasets that can fit into a single V100 GPU.
    \textbf{Overall, \Algo is much better than U-Net and PINN and on par with FNO.} For detailed numerical results, see  Table~\ref{table:pdebench1} in the Appendix. \textbf{Right:}  \Algo  is trained on resolution 256  and directly evaluated on resolution 512. The prediction still matches the ground truth.}
    \label{fig:pde}%
    \vspace{-0.2cm}
  \end{figure*}

After validating \Algo on a broad set of tasks, we dive into two specific modalities, PDE solving and tabular classification,
to show that cross-modal fine-tuning is  promising for model development  in  highly specialized areas. \Algo can  not only achieve high prediction accuracy in both domains, but also recover an important property of Neural Operators---modeling PDEs with zero-shot super-resolution.

\subsubsection*{PDEBench for Scientific ML}
\vspace{-0.1cm}
ML models for physical systems have gained increasing
interest in recent years. To study how cross-modal fine-tuning can help in the scientific ML context, we evaluate \Algo on 8  datasets from PDEBench \citep{Takamoto2022PDEBENCHAE} and compare against
state-of-the-art task-specific models: the physics-informed neural network PINN \citep{Raissi2019PhysicsinformedNN}, Fourier neural operator (FNO) \citep{li2021fno}, and the generic image-to-image regression model U-Net \citep{Ronneberger2015UNetCN}. We focus on the forward  prediction problems. See   Appendix~\ref{appendix:pdebench} for the  experiment details.

\looseness=-1
As shown in Figure~\ref{fig:pde} (left), \textbf{\Algo outperforms PINN and U-Net on all evaluated datasets  and  beats FNO on half of them}, using a smaller training time budget than U-Net and FNO. This is an impressive result given that  the baselines, in particular FNO, are  carefully designed with domain knowledge.  
More crucially, as shown in Figure~\ref{fig:pde} (right), \textbf{\Algo achieves zero-shot super-resolution} (trained on a lower resolution and directly evaluated on a higher resolution)  when using the RoBERTa backbone and an embedder with pointwise convolutions. This generalization ability has only been observed in FNOs.  \Algo also achieves it possibly because the sequence features generated by  pointwise convolutions are  resolution-invariant and can capture the intrinsic flow dynamics. 
These results demonstrate the potential of cross-modal  fine-tuning in the scientific ML context.

\begin{table}[t]
 \caption{\small Tabular results with baselines  from \citet{Hollmann2022TabPFNAT} and \citet{Dinh2022LIFTLF}.   ``Diff. from XGBoost" is the across-task average of per-task difference from XGBoost.   \textbf{\Algo beats  classical approaches and advanced transformer methods on 19 tasks.} For per-task results, see Appendix~\ref{appendix:openml}.}
\label{table:tabsummary}
 \vspace{-0.1cm}
	\centering
 \Large
\resizebox{0.49\textwidth}{!}{	\begin{tabular}{l|ccc|cc|c}
		\toprule
 OpenML-CC18   & LightGBM &        CatBoost &          XGBoost &                        AutoGluon  &           TabPFN   & \Algo \\
		\midrule
\# Wins/Ties    &   1/30	&1/30	&3/30	&\textbf{12/30}	&7/30&	\textbf{12/30}  \\
Avg. AUROC ($\uparrow$) &  0.884  &   0.8898  &   0.8909  &            \textbf{0.8947}   &    0.8943 &    0.8946\\
Diff. from XGBoost &  -6.97E-3 &   -1.18E-3  &   0  &           \textbf{  +3.74E-3 }  &    +3.38E-3 & +3.63E-3\\ 
		\bottomrule
	\end{tabular}}
 \\
  \vspace{0.15cm}
 \Large
\resizebox{0.49\textwidth}{!}{	\begin{tabular}{l|ccc|c|c}
		\toprule
 LIFT Tasks   & LogisticRegression    & SVM    & XGBoost & LIFT GPT-3 & \Algo \\
		\midrule
\# Wins/Ties   &  2/14	&3/14&	2/14&	2/14&	\textbf{7/14} \\
Avg. Acc. ($\uparrow$)
                                         &  79.58 	 &  80.63 &  78.21 	&  79.63 	& \textbf{ 83.80 }\\
    Diff. from XGBoost & +1.37&	+2.42 & 0 &	+1.42&	\textbf{+5.60}\\
		\bottomrule
	\end{tabular}}
	\end{table}

\subsubsection*{OpenML for Tabular Classification} 
Despite being one of the most commonly seen  data types, tabular data are still primarily modeled with classical ML methods like  XGBoost \citep{Chen2016XGBoostAS}. More recently, deep learning approaches such as AutoGluon \citep{Erickson2020AutoGluonTabularRA} and TabPFN \citep{Hollmann2022TabPFNAT} have applied  task-specific transformers  to tabular data with some success.
We next show that \Algo can adapt pretrained RoBERTa to tabular data, outperforming  classical methods and matching the performance of recent deep learning approaches. 

Similar to \citet{Hollmann2022TabPFNAT}, we evaluate \Algo on 30 datasets  from the OpenML-CC18 benchmark \citep{Vanschoren2014OpenMLNS}, comparing against both classical boosting algorithms \citep{Ke2017LightGBMAH, Ostroumova2017CatBoostUB} and advanced transformer-based models \citep{Erickson2020AutoGluonTabularRA, Hollmann2022TabPFNAT}. As shown in Table~\ref{table:tabsummary} (top), \textbf{\Algo ranks  first  on 12/30 tasks} and works as well as AutoGluon, the state-of-the-art AutoML method on tabular data.
It also outperforms TabPFN \citep{Hollmann2022TabPFNAT}, a transformer-based prior-data fitted network, on 16/30 tasks. 

\looseness=-1
It is worth noting that no
single method performs best on all tasks. 
For datasets  where there are limited data  described by categorical variables (e.g., dresses-sales)\footnote{See Table~\ref{table:tabular_results_table} for per-task scores, Table~\ref{table:metadata1} for  task meta-data.}, boosting algorithms perform poorly, but \Algo does significantly better.
For datasets with balanced labels and consisting of a few numerical variables (e.g., diabetes), classical methods are sufficient and less prone to overfitting than large models.
Nonetheless, our results    confirm again that cross-modal fine-tuning can be  appealing  for tackling   real-life problems.

\subsection{Comparison with Task-Specific Cross-Modal Work} 

\looseness=-1
As stated in the introduction, one motivation of \Algo is that the handful of existing cross-modal methods are mostly ad-hoc and tailored to specific modalities. Developing them thus requires a thorough understanding of the  target data. To show that \Algo performs better   while being generally applicable to arbitrary domains, we compare with (1) IGTD \citep{Zhu2021ConvertingTD}, which converts  gene-drug features to images and  applies CNNs to predict drug response; and (2)  LIFT \citep{Dinh2022LIFTLF}, which transforms  tabular data into text to prompt a pretrained GPT-3. 
Table~\ref{table:drug} shows the $R^2$ score for the drug response  tasks, and Table~\ref{table:tabsummary} (bottom) shows the classification accuracy for LIFT datasets. Once again, \textbf{\Algo beats these carefully curated task-specific methods, proving itself as both general and highly effective}. 
\begin{table}
\caption{\small Coefficient of determination  ($R^2$, $\uparrow$) on two drug response prediction datasets. 
 \Algo  outperforms IGTD, which converts raw tabular features to images  to apply vision models.}
 \vspace{-0.2cm}
\label{table:drug}
	\centering
 \small
\resizebox{0.43\textwidth}{!}{\begin{tabular}{lcc}
		\toprule
    $R^2$   & Dataset 1: CTRP   & Dataset 2:  GDSC\\  
		\midrule
	IGTD-CNN &	\res{0.856}{0.003}&\res{0.74}{0.006}\\
	\Algo	  &	\textbf{\res{0.86}{0.002}} &	\textbf{\res{0.831}{0.002}}\\
		\bottomrule
	\end{tabular}}
  \vspace{-0.2cm}
\end{table}

\subsection{Limitation and Future Work}
We identify several future directions based on our experiment results. First, it is worth studying the effect of pretraining modality further and develop a systematic way of selecting pretrained models. Then, we can incorporate model selection into \Algo for a more automated  pipeline. Second, while \Algo leverages the simplest fine-tuning paradigm,
it is possible to combine it with more sophisticated transfer techniques such as adapters \citep{He2022TowardsAU}. {We briefly study how prompting \citep{Bahng2022ExploringVP, Jia2022VisualPT} can be applied to diverse tasks in Appendix~\ref{appendix:prompting} and find that it is  less effective for out-of-modality problems, but we might boost its performance using \Algo.} 
Lastly, we currently evaluate  \Algo on  1D/2D tasks. It is also important to validate it on more  settings, such as high-dimensional problems and reinforcement learning \citep{Reid2022CanWH}.

\section{Conclusion}
In this paper, we study how we can reuse existing models  for new and less-explored areas. We propose a novel and effective cross-modal fine-tuning framework, \Algo, that aligns the end-task data from an arbitrary modality with a model's pretraining modality to improve fine-tuning performance. 
Our work not only signals the potential of large-scale pretraining for diverse tasks but also lays out a path for a largely uncharted data-centric paradigm in ML.

\

%
\vspace{-0.5cm}
\section*{Acknowledgments}
We thank Noah Hollmann  for providing useful feedback on the tabular experiments. This work was supported in part by the National Science Foundation grants IIS1705121, IIS1838017, IIS2046613, IIS2112471, and funding from Meta, Morgan Stanley, Amazon, and Google. Any opinions, findings and conclusions or recommendations expressed in this material are those of the author(s) and do not necessarily reflect the views of any of these funding agencies.

\nocite{langley00}

\bibliography{reference}
\bibliographystyle{icml2023}

\newpage
\appendix
\onecolumn
\newpage
\appendix
\section{Appendix}

\subsection{Embedding Learning with Optimal Transport Dataset Distance }
\label{appendix:otdd}

\subsubsection{Literature Review}
Due to the limited space, we do not give a full review of the optimal transport dataset distance (OTDD) \citep{AlvarezMelis2020GeometricDD} in the main text. Here, we briefly recall the optimal transport (OT) distance and explain OTDD in detail.

Consider a complete and separable metric space $\mathcal{X}$ and let $\mathcal{P}(\mathcal{X})$ be the set of probability measures on $\mathcal{X}$. 
For $\alpha, \beta \in \mathcal{P}(\mathcal{X})$, let $\Pi(\alpha, \beta)$ be the set of joint probability distributions on $\mathcal{X} \times \mathcal{X}$ with marginals $\alpha$ and $\beta$ in the first and second dimensions respectively.  Then given a cost function $c(\cdot, \cdot): \mathcal{X} \times \mathcal{X} \rightarrow \mathbb{R}^+$, the classic OT distance with cost $c$ is defined by:
\begin{align}
    \OT_c(\alpha, \beta) := \min_{\pi \in \Pi(\alpha, \beta)} \int_{\mathcal{X} \times \mathcal{X}} c(x, y) d\pi(x, y).
\label{equ:ot}
\end{align}
When $\mathcal{X}$ is equipped with a metric $d_\mathcal{X}$, we can use $c(x, y) = d_X(x, y)^p$ for some $p\geq 1$ and obtain the $p$-Wasserstein distance, $W_p(\alpha, \beta) := (\OT_{d_\mathcal{X}^p}(\alpha, \beta))^\frac{1}{p}$. 

Now consider the case of finite datasets with features in $\mathcal{X}$ and labels in a finite set $\mathcal{Y}$.  Each dataset can be considered a discrete distribution in $\mathcal{P}(\mathcal{X} \times \mathcal{Y})$.  
To define a distance between datasets, a natural approach is to define an appropriate cost function on $\mathcal{Z} := \mathcal{X} \times \mathcal{Y}$ and consider the optimal transport distance.  Indeed, for any metric $d_\mathcal{Y}$ on $\mathcal{Y}$ and any $p \geq 1$,  $\mathcal{Z}$ can be made a complete and separable metric space with metric
\begin{align}
    d_\mathcal{Z} ((x, y), (x', y')) = (d_\mathcal{X} (x, x')^p + d_\mathcal{Y}(y, y')^p)^\frac{1}{p}
    \label{equ:dz}
\end{align}

It is usually not clear how to define a natural distance metric in $\mathcal{Y}$, so instead we proceed by representing each class $y \in \mathcal{Y}$ by $P(\mathcal{X} | \mathcal{Y} = y)$, the conditional distribution of features $\mathcal{X} $ given $\mathcal{Y}=y$. 
More specifically, for a dataset $\mathcal{D} \in \mathcal{P}(\mathcal{X} \times \mathcal{Y})$, denote this map from classes to conditional distributions by $F(\mathcal{D}, \cdot): \mathcal{Y} \to \mathcal{P}(\mathcal{X})$.  Then we can transform any dataset over $\mathcal{X} \times \mathcal{Y}$ into one over $\mathcal{X} \times \mathcal{P}(\mathcal{X})$ via $G(\mathcal{D}) := (\text{proj}_\mathcal{X}, F(\mathcal{D}, \text{proj}_Y))$.

As discussed above, $W_p$ is a natural notion of distance in $\mathcal{P}(\mathcal{X})$, so by substituting $\mathcal{Y} \mapsto \mathcal{P}(\mathcal{X})$ and $d_\mathcal{Y} \mapsto W_p$ in Equation \ref{equ:dz}, we can define the ($p$-)optimal transport dataset distance between datasets $\mathcal{D}_A$ and $\mathcal{D}_B$ by
\begin{align}
    \OTDD(\mathcal{D}_A, \mathcal{D}_B) := \OT_{(d_\mathcal{X}^p \times W_p^p)^\frac{1}{p}}(G(\mathcal{D}_A), G(\mathcal{D}_B))
\end{align}

\subsubsection{Computational considerations}
As we aim for a practical fine-tuning workflow, computational cost is a crucial concern. While \citet{AlvarezMelis2020GeometricDD} proposed two variants of OTDD---the exact one and a Gaussian approximation, we observe from our experiments that optimizing the exact OTDD leads to better performance. In the following, we will focus on analyzing the computational cost of the exact OTDD.

Given datasets with $D$-dimensional feature vectors, estimating vanilla OT distances can be computationally expensive and has a worst-case complexity of $O(D^3 \log D)$ \citep{Pele2009FastAR}. However,  adding an entropy regularization term $\epsilon H(\pi | \alpha \otimes \beta)$ to Equation~\ref{equ:ot}, where $H$ is the relative entropy and $\epsilon$ controls the time-accuracy trade-off, can be solved efficiently with the Sinkhorn algorithm \citep{Cuturi2013SinkhornDL}. This reduces OT's empirical complexity to $O(D^2)$ and makes the time cost for computing OTDD manageable for 
\Algo's workflow.

\begin{algorithm}[tb]
  \caption{Efficient approximation of OTDD using class-wise subsampling.}
  \label{alg:}
\begin{algorithmic}
  \STATE {\bfseries Input:} target dataset $\{x^t, y^t\}$, number of target classes $K^t$, source dataset $S=\{x^s, y^s\}$, subsample size $b$, subsample round $R$
  \FOR{each class $i \in [K^t]$ in the target dataset}
  \STATE Compute class weight $w_i = \frac{\text{number of target data in class } i}{\text{total number of target data}}$
  \STATE Generate data loader $D_i$ consisting of data in class $i$
  \ENDFOR
  \FOR{$i \in [K^t]$} 
  \FOR{$r \in [R]$}
  \STATE Subsample $b$ target data points $D_{ir}$ uniformly at random from $D_i$
  \STATE Compute class-wise distance $d_{ir} = OTDD(D_{ir},S)$
  \ENDFOR
  \STATE Approximate class-wise OTDD by $d_i = \frac{1}{R}\sum_{i=1}^{R} d_{ir}$
  \ENDFOR
  \STATE Approximate OTDD by $d = \sum_{i=1}^{K^t} w_i \cdot d_i$
\end{algorithmic}
\end{algorithm}

\begin{figure}[t!]
  \begin{center}
    \includegraphics[width=0.7\textwidth]{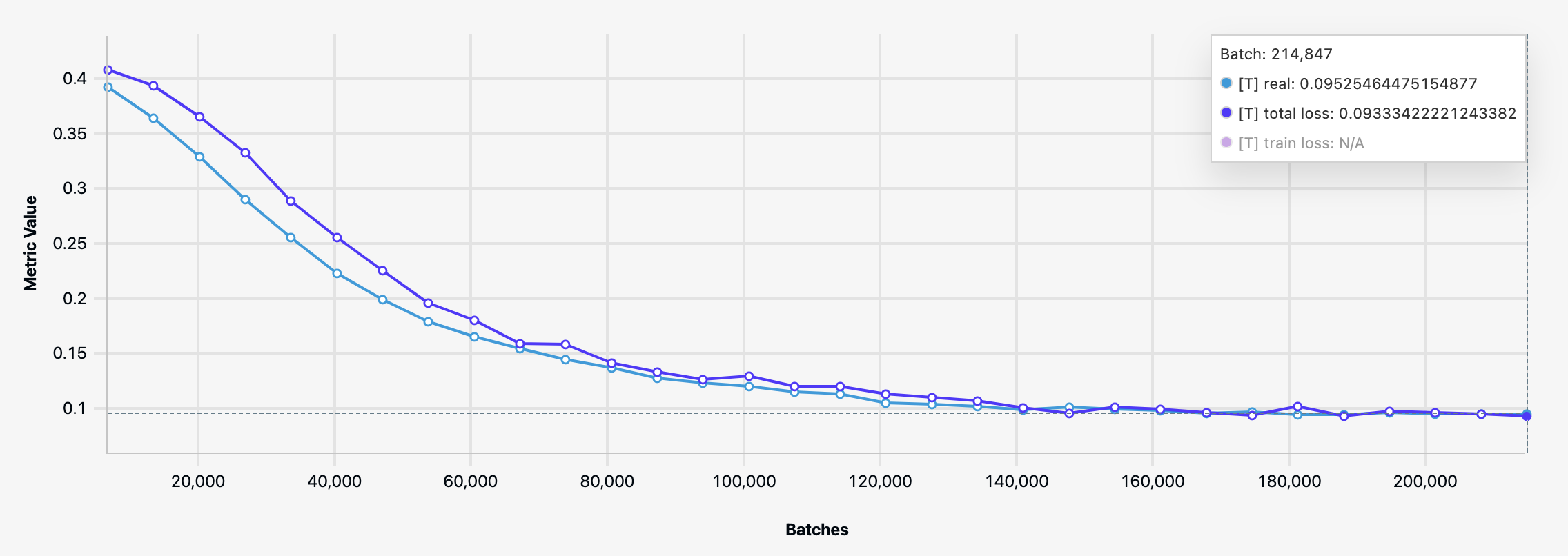}
  \end{center}
  \vspace{-0.2cm}
  \caption{Screenshot of  OTDD curves during embedding learning in one task. x-axis is the number of optimization steps, y-axis represents OTDD (1E2).  We use Algorithm~\ref{alg:} to approximate the exact OTDD as the loss function for optimization on GPU (purple curve). We also  track the actual OTDD on CPU (blue curve). We can see that the proposed algorithm works well, which allows us to perform embedding learning efficiently.  
  }
  \label{fig:algo}
  \vspace{-0.3cm}
\end{figure}
During implementation of \Algo, we also observed memory issues for computing OTDD using the entire target and source datasets on GPUs. To alleviate this, we reduce the dimensionality of the feature vectors by taking the average along the sequence length dimension. We further propose a class-wise subsampling strategy for approximating OTDD on GPUs (Algorithm~\ref{alg:}). In short, we split the $K$-class target dataset into $K$ datasets based on the labels and compute the class-wise OTDD between each single-class target dataset and the \textit{entire source dataset}. Each class-wise OTDD can be approximated with the average of batch samples similar to how stochastic gradient descent approximates gradient descent. After that, we approximate the OTDD between the target and source datasets using the weighted sum of the $K$ class-wise OTDDs. To verify that the approximation works empirically, we track the approximated OTDD (computed on GPUs) and the actual OTDD (computed on CPUs) and visualize the loss curves during \Algo's embedder learning process (Figure~\ref{fig:algo}). We can see that the estimated value adheres to the actual value. 

Leveraging both the Sinkhorn algorithm and class-wise approximation, the embedder learning process only takes up a small fraction of the total fine-tuning time in practice, as shown in Table~\ref{table:timebreakdown} in the later experiment results section. Hence, we invest a reasonable time budget but achieve significantly improved cross-domain transfer performance using \Algo.

\subsection{\Algo Implementation}
\label{appendix:implementation}
\subsubsection{Pretrained Models}
We evaluated \Algo with two pretrained models in our experiments. In Table~\ref{table:accwitherror}, for all 2D tasks including CIFAR-100, Spherical, Darcy Flow, PSICOV, Cosmic, NinaPro, and FSD50K, we use the following model. As Swin has a pretrained resolution, we reshape the inputs for our tasks to the resolution before feeding them into the model.
\begin{table}[h!]
\centering
\begin{tabular}{cccccc}
			\toprule
			 Name & Pretrain &	Resolution & Num Params & FLOPS &FPS  \\
			\midrule
			Swin-base \citep{Liu2021SwinTH} &ImageNet-22K&224$\times$224&88M& 15.4G& 278 \\
			\bottomrule
		\end{tabular}
\label{table:swin}
\end{table}

For all 1D tasks including ECG, Satellite, DeepSEA, JSB Chorales, ListOps,and Homology, we use the following model:
\begin{table}[h!]
\centering
\begin{tabular}{ccccc}
			\toprule
			 Name & Pretrain  & Num Params & FLOPS   \\
			\midrule
			RoBERTa-base \citep{Liu2019RoBERTaAR} &Five English-language corpora&125M& 1.64E20 \\
			\bottomrule
		\end{tabular}
\label{table:roberta}
\end{table}

We use the Hugging Face transformers library \cite{Wolf2019HuggingFacesTS} to implement the pretrained models.

\subsubsection{Hyperparameter Tuning}
\label{appendix:hptuning}
As \Algo is both task-agnostic and model-agnostic, it can be applied to fine-tuning a variety of pretrained transformers on drastically different end tasks with distinct datasets. Hence, it is hard to define one set of fine-tuning hyperparameters for all (model, task) pairs. At the same time, optimizing large-scale pretrained transformers can be challenging due to their large model sizes, as the downstream performance depends largely on the hyperparameters used. For instance, using a large learning rate can  distort pretrained weights and lead to catastrophic forgetting. Therefore, in our experiments, given a (model, task) pair,  we first apply hyperparameter tuning using the Asynchronous Successive Halving Algorithm (ASHA) \citep{li2020system} to the \textit{standard fine-tuning setting} (i.e., after initializing the embedder and predictor architectures, directly updating all model weights to minimize the task loss) to identify a proper training configuration. Then, we use the same set of hyperparameters found for all our experiments for the particular (model, task) combination. Note that even though we did not explicitly state this in the main text, the hyperparameter tuning stage can be directly integrated into the \Algo workflow between stage 1 and stage 2. In this sense, \Algo is still an automated cross-modal transfer workflow that works for diverse tasks and different pretrained models.

The configuration space for ASHA can be customized for each task. In general, the following search space is sufficient:
\begin{itemize}
\item Target sequence length:  8, 64, 512 for RoBERTa
\item Batch size: 4, 16, 64
\item Gradient clipping: -1, 1
\item Dropout: 0, 0.05
\item Optimizer: SGD, Adam, AdamW
    \item Learning rate: 1E-2, 1E-3, 1E-4, 1E-5
    \item Weight decay: 0, 1E-2, 1E-4
\end{itemize}

\subsubsection{More Details on Embedder Architecture Design}
In the current  workflow, we use the following procedure to determine the kernel size $k$ for the embedder's convolution layer:

\begin{itemize}
    \item For RoBERTa: we apply hyperparameter search to the vanilla fine-tuning baseline to find the optimal sequence length $s^*$ for the second dimension of the embedder output with shape  $(batch\_size, seq\_len, embed\_dim)$. The configuration space is $\{8, 64, 512\}$. Then, $k$ is set to largest value such that after applying convolution with $c_{out}= embed\_dim$ (e.g., $768$ for RoBERTa) and transposing the last two dimensions, the $seq\_len$ dimension of the output tensor is closest to the searched value $s^*$. For example, if the input length is $1024$ and the searched $s^*$ is $256$, then $k$ (and the stride) is $4$, so the output of the conv layer has shape $(batch\_size, 768, 256)$. We then transpose it to get $(batch\_size, 256, 768)$.
    \item For Swin: given that Swin Transformers already have the patchify operation, we want to reuse the pretrained patchify layer, which has $k=4$. Thus, given the target task, we first resize the height and width of the target input to those of the pretraining data, e.g.,  $(224, 224)$ for models pretrained with ImageNet. Then, the pretrained patchify layer with $k=4$ can be reused by the embedder.
\end{itemize}
\subsubsection{Embedding learning with OTDD}
After initializing the embedder architecture for each task, we train it to minimize the OTDD between the embedded target features and embedded source features. 

For source datasets, we use CIFAR-10 for Swin and CONLL-2003 for RoBERTa. We sample 5000 data points to compute OTDD. In practice, we can pass the source data through the pretrained embedder once and save all the embedded features, so we don't have to pay the cost of obtaining the source features each time we fine-tune a new model.

For classification tasks, we directly use the labels provided by the end task to compute OTDD. For dense tasks, we perform K-Means clustering on the target data to obtain pseudolabels for OTDD computation. The number of clusters is set to the number of classes of the source dataset, e.g., 10 for 2D tasks that use CIFAR-10 as the source dataset.

To compute the embedding learning objective, we use the OTDD implementation of the original paper provided here: \url{https://github.com/microsoft/otdd}. We use the searched hyperparameters in Section~\ref{appendix:hptuning}. The others are fixed across different tasks:
\begin{itemize}
    \item Embedding learning epochs: 60
    \item Embedding learning stage rate scheduler:  decay by 0.2 every 20 epochs
    \item Fine-tuning stage learning rate scheduler: we use the linear decay with min\_lr = 0 and 5 warmup epochs
\end{itemize}

\subsection{Baseline Implementation}
For the standard fine-tuning baseline, we use the same hyperparameter configuration (number of epochs, batch size, learning rate, etc) as \Algo, except for setting embedding learning epochs to 0.

For the train-from-scratch baseline, everything is the same as standard fine-tuning, except that the model weights are reinitialized at the beginning.

\subsection{Experiments on NAS-Bench-360}
\label{appendix:experiment}
\subsubsection{Information About the Benchmark and Experiment Protocol}
\label{appendix:taskinfo}
\begin{table}[h!]
	\vspace{0.3cm}
 \caption{Summary about each  task and the hand-designed expert models used
 in NAS-Bench-360 \citep{nasbench360}.}
\label{table:taskinfo}
	\centering
	\resizebox{1\textwidth}{!}{\begin{tabular}{lllllll}
		\toprule
		Task name   & \# Data  & Data dim. & Type & License &  Learning objective &Expert arch.\\
		\midrule
		\multirow{2}{*}{CIFAR-100} & \multirow{2}{*}{60K}  & \multirow{2}{*}{2D} & \multirow{2}{*}{Point} & \multirow{2}{*}{CC BY 4.0}& \multirow{2}{*}{Classify natural images into 100 classes} & DenseNet-BC \\
		&&& && & \citep{Huang2017DenselyCC}\\
		\midrule
		\multirow{2}{*}{Spherical} & \multirow{2}{*}{60K}  &  \multirow{2}{*}{2D} & \multirow{2}{*}{Point} &\multirow{2}{*}{CC BY-SA}&  Classify spherically projected images   &S2CN \\
&& &&& into 100 classes &\citep{Cohen2018SphericalC}\\
		\midrule
		\multirow{2}{*}{NinaPro} & \multirow{2}{*}{3956}  &  \multirow{2}{*}{2D}& \multirow{2}{*}{Point} &\multirow{2}{*}{CC BY-ND}&  Classify sEMG signals into 18 classes   & Attention Model\\
& &&&&corresponding to hand gestures &\citep{Josephs2020sEMGGR} \\
		\midrule
		\multirow{2}{*}{FSD50K} & \multirow{2}{*}{51K} &  \multirow{2}{*}{2D} & 
		Point & \multirow{2}{*}{CC BY 4.0}&Classify sound events in log-mel & VGG\\ 
& &&(multi-label)&  &spectrograms with 200 labels &\citep{Fonseca2021FSD50KAO} \\
		\midrule
		\multirow{2}{*}{Darcy Flow} & \multirow{2}{*}{1100}  & \multirow{2}{*} {2D} & \multirow{2}{*}{Dense} & \multirow{2}{*}{MIT}& Predict the final state of a fluid from its  & FNO\\
&& &&& initial conditions &\cite{li2021fno}\\	
		\midrule
		\multirow{2}{*}{PSICOV} & \multirow{2}{*}{3606} &  \multirow{2}{*}{2D}  & \multirow{2}{*}{Dense} &\multirow{2}{*}{GPL}&  Predict pairwise distances between resi- &DEEPCON \\
&&& && duals from 2D protein sequence features & \citep{10.1093/bioinformatics/btz593}\\

		\midrule
\multirow{2}{*}{Cosmic} & \multirow{2}{*}{5250} &  \multirow{2}{*}{2D}  & \multirow{2}{*}{Dense} & \multirow{2}{*}{Open License}& Predict propablistic maps to identify cos- &deepCR-mask \\
&& &&& mic rays in telescope images  &\citep{Zhang2019deepCRCR} \\	

		\midrule 
\multirow{2}{*}{ECG} & \multirow{2}{*}{330K} &  \multirow{2}{*}{1D}  & \multirow{2}{*}{Point} & \multirow{2}{*}{ODC-BY 1.0} &  Detect atrial cardiac disease from & ResNet-1D\\
&& &&& a ECG recording (4 classes) &\citep{Hong2020HOLMESHO} \\

		\midrule 
\multirow{2}{*}{Satellite} & \multirow{2}{*}{1M} &  \multirow{2}{*}{1D}  & \multirow{2}{*}{Point}&
\multirow{2}{*}{GPL 3.0}  &  Classify satellite image pixels' time   & ROCKET \\
&& &&&  series into 24 land cover types  &\citep{Dempster2020ROCKETEF} \\

		\midrule
\multirow{2}{*}{DeepSEA} & \multirow{2}{*}{250K} &  \multirow{2}{*}{1D} & 
Point & \multirow{2}{*}{CC BY 4.0} &Predict chromatin states and binding  &
\multirow{2}{*}{} DeepSEA\\ 
&&&(multi-label)&  &states of RNA sequences (36 classes)  &\citep{Zhou2015PredictingEO} \\
		\bottomrule
	\end{tabular}}
\vspace{0.3cm}
\end{table}
For experiments, each dataset is preprocessed and split using the script available on \url{https://github.com/rtu715/NAS-Bench-360}, with the training set being used for hyperparameter tuning, embedding learning, and fine-tuning.

When training/fine-tuning is finished, we evaluate the performance of all models following the NAS-Bench-360 protocol. We first report results of the target metric for each task by running the model of the \textit{last} epoch on the test data. Then, we report aggregate results via performance profiles~\citep{dolan2002profiles}, a technique that considers both outliers and small performance differences to compare methods across multiple tasks robustly. In such plots, each curve represents one method. The $\tau$ on the $x$-axis denotes the fraction of tasks on which a method is no worse than a $\tau$-factor from the best. The performance profile for our experiments is shown in Figure~\ref{fig:profile}.

The code and  configuration file for reproducing each experiment can be found in our \href{https://github.com/sjunhongshen/ORCA}{official GitHub repository}.

\subsubsection{Complete Results for Table~\ref{table:accwitherror} with Error Bars}
\label{appendix:nb360std}

\begin{table*}[h]
\vspace{0.3cm}
\caption{Prediction errors ($\downarrow$) for  10 diverse tasks. 	``NAS-Bench-360" refers to the task-wise best of all AutoML baselines evaluated in the paper, including DARTS~\citep{liu2018darts}, DenseNAS~\citep{Fang2020DenselyCS},   AMBER~\citep{Zhang2020AnAF}, Auto-DL~\citep{Liu2019AutoDeepLabHN},  WRN-ASHA~\citep{li2020system}, and XGBoost~\citep{Chen2016XGBoostAS}. ``FPT" refers to fine-tuning the layer norms of RoBERTa/Swin. On 7/10 problems, \Algo ranks the first among all competitors. 
		}
		\large
\resizebox{1.0\textwidth}{!}{	\begin{tabular}{lcccccccccc}
			\toprule
			 & CIFAR-100 & Spherical  & Darcy Flow & PSICOV & Cosmic& NinaPro & FSD50K  & ECG & Satellite & DeepSEA  \\
			  & 0-1 error (\%)  & 0-1 error (\%) & relative $\ell_2$ & MAE$_8$  & 1-AUROC   & 0-1 error (\%)   & 1- mAP & 1 - F1 score  & 0-1 error (\%)  & 1- AUROC \\
			
			\midrule
			Hand-designed  &\res{19.39}{0.20}  & \res{67.41}{0.76}  &\res{8E-3}{1E-3} & \res{3.35}{0.14} &  \textbf{\res{0.127}{0.01}} &\res{8.73}{0.90}  & \res{0.62}{0.004} & \textbf{\res{0.28}{0.00} } & \res{19.80}{0.00} & \res{0.30}{0.024} \\

            \midrule
		  NAS-Bench-360&\res{23.39}{0.01}  & \res{48.23}{2.87} & \res{2.6E-2}{1E-3}  & \res{2.94}{0.13} & \res{0.229}{0.04} & \res{7.34}{0.76} & \res{0.60}{0.001} &\res{0.34}{0.01}& \res{12.51}{0.24} &  \res{0.32}{0.010} \\ 
			DASH &\res{24.37}{0.81}  & \res{71.28}{0.68}  & \res{7.9E-3}{2E-3}  & \res{3.30}{0.16} &\res{0.19}{0.02} &\textbf{\res{6.60}{0.33}} & \res{0.60}{0.008}  & \res{0.32}{0.007}  & \res{12.28}{0.5} & \textbf{\res{0.28}{0.013}}\\
			\midrule 	
			Perceiver IO  & \res{70.04}{0.44} & \res{82.57}{0.19} & \res{2.4E-2}{1E-2} & \res{8.06}{0.06} & \res{0.485}{0.01} & \res{22.22}{1.80} & \res{0.72}{0.002} &\res{0.66}{0.01}& \res{15.93}{0.08} & \res{0.38}{0.004} \\
			FPT &\res{10.11}{1.18} & \res{76.38}{4.89}&\res{2.1E-2}{1.3E-3} & \res{4.66}{0.054} &\res{0.23}{0.002}& \res{15.69}{2.33}&\res{0.67}{0.0068} &\res{0.50}{0.0098} & \res{20.83}{0.24}&\res{0.37}{0.0002} \\
			\midrule
		\textbf{\Algo}  &\textbf{\res{6.53}{0.079}}  & \textbf{\res{29.85}{0.72}}&\textbf{\res{7.3E-3}{6.8E-5}} & \textbf{\res{1.91}{0.038}} &\res{0.152}{0.005}& \res{7.54}{0.39}&\textbf{\res{0.56}{0.013}} &\textbf{\res{0.28}{0.0059}}&\textbf{ \res{11.59}{0.18}}  & \res{0.29}{0.006}  \\
			\bottomrule
		\end{tabular}}
  		
	\end{table*}

\subsubsection{Complete Results for Table~\ref{table:layernorm} with Error Bars}
\label{appendix:layernormstd}
\begin{table*}[h]
		\vspace{0.3cm}
  \caption{Prediction errors ($\downarrow$) of \Algo, vanilla fine-tuning, and training RoBERTa/Swin from scratch. We consider  fine-tuning all parameters (full setting) vs. only the layer norms (FPT setting). \Algo is better in both  settings.
		}
		\large
\resizebox{1.0\textwidth}{!}{		\begin{tabular}{lcccccccccc}
			\toprule
			 & CIFAR-100 & Spherical  & Darcy Flow & PSICOV & Cosmic& NinaPro & FSD50K & ECG & Satellite & DeepSEA  \\	
			\midrule
			Train-from-scratch &\res{50.87}{0.32} & \res{76.67}{0.21}&\res{8.0E-2}{1.3E-2} &
			\res{5.09}{0.014}&\res{0.50}{0.00}&
			\res{9.96}{1.67}&\res{0.75}{0.017}&\res{0.42}{0.011} & \res{12.38}{0.14} & \res{0.39}{0.01}\\
				\midrule
				Fine-tuning &\res{7.67}{0.55} & \res{55.26}{1.63}&\res{7.34E-3}{1.1E-4} & \res{1.92}{0.039}&\res{0.17}{0.011}& \res{8.35}{0.75}&\res{0.63}{0.014} &\res{0.44}{0.0056} & \res{13.86}{1.47}&\res{0.51}{0.0001} \\
		\textbf{\Algo} &\textbf{\res{6.53}{0.079}}  & \textbf{\res{29.85}{0.72}}&\textbf{\res{7.28E-3}{6.8E-5}} & \textbf{\res{1.91}{0.038}} &\textbf{\res{0.152}{0.005}}&\textbf{ \res{7.54}{0.39}}&\textbf{\res{0.56}{0.013}} &\textbf{\res{0.28}{0.0059}}&\textbf{ \res{11.59}{0.18}}  & \textbf{\res{0.29}{0.006}}\\
				\midrule
			Fine-tuning (layernorm) &\res{10.11}{1.18} & \res{76.38}{4.89}&\res{2.1E-2}{1.3E-3} & \res{4.66}{0.054} &\res{0.233}{0.002}& \res{15.69}{2.33}&\res{0.67}{0.0068} &\res{0.50}{0.0098} & \res{20.83}{0.24}&\res{0.37}{0.0002}\\
		\textbf{\Algo (layernorm)} &\res{7.99}{0.098} & \res{42.45}{0.21}&\res{2.1E-2}{7.4E-4} & \res{4.97}{0.14} &\res{0.227}{0.003}& \res{15.99}{1.92}&\res{0.64}{0.0093}&\res{0.47}{0.007} & \res{20.54}{0.49} & \res{0.36}{0.0070}\\
			
			\bottomrule
		\end{tabular}}
  
	\end{table*}

\subsubsection{Ablation Study on Embedding Learning Metrics}
\label{appendix:metrics}

As motivated in Section~\ref{sec:exp:finetuneablation}, we present here an ablation study on the embedding learning metrics that we have considered for minimizing distribution dissimilarity. The results show that (1) performing feature alignment generally helps downstream adaptation, regardless of which metric we minimize; (2) OTDD leads to the best overall performance, so we chose it for our workflow. Our findings confirm that it is the general idea of data alignment, rather than a specific metric, that makes cross-modal transfer work.

Specifically, we experiment with OTDD, maximum mean discrepancy (MMD) \citep{Gretton2012AKT}, and pairwise Euclidean distance. We learn the embedders to minimize these metrics and then fine-tune the pretrained models. The test errors are as follows, which are used to plot  the performance profiles in Figure~\ref{fig:profile} (right).

\begin{table}[h!]
		\centering
  \vspace{0.3cm}
  \caption{Prediction errors ($\downarrow$) of different distance metrics. OTDD achieves the best overall performance. ``Naive fine-tuning" represents fine-tuning without embedder learning.
		}
		\label{table:metric}
\resizebox{1\textwidth}{!}{		\begin{tabular}{lcccccccccc}
			\toprule
			 & CIFAR-100 & Spherical  & Darcy Flow & PSICOV & Cosmic &	NinaPro & FSD50K  & ECG & Satellite & DeepSEA \\
	
			\midrule
			OTDD &\textbf{\res{6.53}{0.079}}  & \textbf{\res{29.85}{0.72}}&\textbf{\res{7.28E-3}{6.8E-5}} & \res{1.91}{0.038} &\textbf{\res{0.152}{0.005}}& \res{7.54}{0.39}&\textbf{\res{0.56}{0.013}} &\textbf{\res{0.28}{0.0059}}& \res{11.59}{0.18}  & \textbf{\res{0.29}{0.006}} \\
			MMD&\res{6.62}{0.092} & \res{33.64}{2.57}&\res{7.4E-3}{3.4E-4} &
		\textbf{	\res{1.9}{0.016}}&\res{0.156}{0.002}&
		\textbf{	\res{7.48}{0.23}}&\res{0.58}{0.004}&\res{0.40}{0.018} & \textbf{\res{11.29}{0.087}} & \res{0.38}{0.077} \\
		Euclidean &\res{7.09}{0.48} & \res{32.33}{2.03}&{\res{7.3E-3}{1.9E-4}} & \res{1.91}{0.019} &\res{0.157}{0.002}& \res{7.51}{0.11}&\res{0.59}{0.02}&\res{0.41}{0.009} & \res{11.4}{0.078} & \res{0.34}{0.002} \\
	\midrule
		Naive fine-tuning &\res{7.67}{0.55} & \res{55.26}{1.63}& \res{7.3E-3}{1.1E-4} & \res{1.92}{0.039}&\res{0.174}{0.011}& \res{8.35}{0.75}&\res{0.63}{0.014} &\res{0.44}{0.0056} & \res{13.86}{1.47}&\res{0.51}{0.0001}\\
		
			\bottomrule
		\end{tabular}}
  
	\end{table}

\newpage
\subsubsection{Ablation Study on Layernorm Initialization}
As discussed in Section~\ref{sec:architecturedesign}, our embedder architecture contains a  layernorm layer. For \Algo, we warm initialize the parameters of  layernorm with those of the pretrained model.  To see how this initialization strategy affects the performance, we additionally evaluate standard fine-tuning with warm initializing the layernorms. As shown in the table below, the effect of warm initialization is task-dependent, i.e, it helps adaptation for  tasks like Spherical and Cosmic but slightly hurts the performance for tasks like Darcy Flow.
\label{appendix:init}
\begin{table*}[h]
  \caption{Prediction errors ($\downarrow$) of \Algo, vanilla fine-tuning, and fine-tuning with warm initializing the layernorm.
		}
		\large
\resizebox{1.0\textwidth}{!}{		\begin{tabular}{lcccccccccc}
			\toprule
			 & CIFAR-100 & Spherical  & Darcy Flow & PSICOV & Cosmic& NinaPro & FSD50K & ECG & Satellite & DeepSEA  \\	
			\midrule
	\textbf{\Algo} &\textbf{\res{6.53}{0.079}}  & \textbf{\res{29.85}{0.72}}&\textbf{\res{7.28E-3}{6.8E-5}} & \textbf{\res{1.91}{0.038}} &\textbf{\res{0.152}{0.005}}&\textbf{ \res{7.54}{0.39}}&\textbf{\res{0.56}{0.013}} &\textbf{\res{0.28}{0.0059}}&\textbf{ \res{11.59}{0.18}}  & \textbf{\res{0.29}{0.006}}\\
			
				Fine-tuning &\res{7.67}{0.55} & \res{55.26}{1.63}&\res{7.34E-3}{1.1E-4} & \res{1.92}{0.039}&\res{0.17}{0.011}& \res{8.35}{0.75}&\res{0.63}{0.014} &\res{0.44}{0.0056} & \res{13.86}{1.47}&\res{0.51}{0.0001} \\

			Fine-tuning (warm init) &\res{6.87}{0.038} &\res{32.51}{1.48}&\res{7.98E-3}{7.18E-5}&\res{2.04}{0.0077}&\res{0.163}{0.003}&\res{9.56}{0.26}&\res{0.62}{0.006}&\res{0.30}{0.011}&\res{12.49}{0.04}&\res{0.33}{0.006}\\

			\bottomrule
		\end{tabular}}
  
	\end{table*}

\subsubsection{Runtime of \Algo vs. FPT}
\label{appendix:runtimefpt}

We record the time for each stage of \Algo in Table~\ref{table:timebreakdown}. We can see that the embedder learning process only takes up a small fraction of the total fine-tuning time in practice

\begin{table}[h!]
\vspace{0.3cm}
 \caption{We record the runtime (in hours) of \Algo's embedding learning stage and the fine-tuning stage for each task. Then, we compute the ratio between the two. Averaged across tasks, embedding learning with OTDD only takes about 11\% of the time needed for fine-tuning. All experiments are performed on NVIDIA V100 GPUs.}
\label{table:timebreakdown}
	\centering
\resizebox{1.0\textwidth}{!}{	\begin{tabular}{lcccccccccc}
		\toprule
		
		     & CIFAR-100 & Spherical  & Darcy Flow & PSICOV & Cosmic & NinaPro & FSD50K &   ECG & Satellite & DeepSEA \\ 
		\midrule
		Embedding & 1.6 & 1.8& 0.18& 0.28&0.25&0.3&0.21 & 0.69 & 0.26&0.2\\
		\midrule 
		Fine-tuning  & 9.2 & 9.3& 0.86& 3.47&2.95&1.1&12.5 & 10.1 & 37.5& 7.6\\
		\midrule
		$\frac{\text{Embedding}}{\text{Fine-tuning}}$ &17\%  &19\%& 20\% &8\%&8\%&27\%&2\%& 7\% &1\% & 3\%\\ 
		\bottomrule
	\end{tabular}}

\end{table}

In Table~\ref{table:layernorm}, we also compare with the FPT setting, which only fine-tunes the layer norms of the pretrained transformer models. As we have shown already, the downstream performance of fine-tuning only a subset of the parameters is less competitive than fine-tuning all parameters. Below, we show that the time saved for updating only layer norms is also not that significant. Therefore, we suggest performing full fine-tuning when time and computational resources allow. 

\begin{table}[h!]
\vspace{0.3cm}
\caption{We record the total runtime (in hours) for four settings: \Algo with full fine-tuning, \Algo with tuning layer norms,  full fine-tuning (without embedding learning), and fine-tuning layer norms (FPT). We can see that tuning the layer norms does not bring significant benefit in terms of reducing  the model development time, but it sacrifices the downstream performance of the resulting models.}
\label{table:timevsfpt}
	\centering
\resizebox{1.0\textwidth}{!}{	\begin{tabular}{lllllllllll}
		\toprule
		     & CIFAR-100 & Spherical  & Darcy Flow & PSICOV & Cosmic & NinaPro & FSD50K &   ECG & Satellite & DeepSEA \\ 
		\midrule
		\Algo & 10.8 & 11.1& 1.04& 3.75&3.2&1.4&12.71 & 10.79 & 37.76&7.8\\
		\Algo (layernorm) & 8.7 & 8.9& 0.76 &3.35&3.1&1.0&8.96& 9.05 &25.56& 5.7\\ 
		\midrule
		Fine-tuning  & 9.2 & 9.3& 0.86& 3.4&2.7&1.1&12.5 & 10.2 & 37.5& 7.4\\
		Fine-tuning (layernorm) & 7.1 & 7.1& 0.58 &3.1&2.5&0.7&8.75& 8.5 & 25.3& 5.5\\ 
		\bottomrule
	\end{tabular}}
 
\end{table}

\newpage
\subsubsection{Results for Applying Different Model Bodies to DeepSEA and Spherical}
	\begin{table}[h]
 	\caption{Prediction errors and post-alignment OTDDs for different pretrained  model bodies. Smaller OTDD leads to smaller  errors.
		 }
		\label{table:modality}
		\centering
		\large
		\resizebox{0.5\textwidth}{!}{\begin{tabular}{lcccc}
			\toprule
			Error (OTDD)     &  DeepSEA (1D)  &  Spherical (2D) \\ 
			\midrule
			RoBERTa (1D) &  \textbf{\res{0.295}{0.006} (37.40)}  &  \res{68.28}{0.017} (19.54)  \\
			\midrule 
			Swin (2D)  & \res{0.361}{0.001} (64.83)& \textbf{\res{29.85}{0.072}(11.78)} \\
			\bottomrule
		\end{tabular}}
	\end{table}
 
\subsection{Experiments on PDEBench}
\label{appendix:pdebench}
We test \Algo on all datasets in PDEBench except for 2D and 3D Navier-Stokes, which could not fit into the memory of a single V100 GPU. For each data, we select one set of parameters and initial conditions, as described in Table~\ref{table:pdebench1}.
We follow the \href{https://github.com/pdebench/PDEBench}{official GitHub repo} of PDEBench to download, preprcoess, and load the data. We use the normalized RMSE, which is scale-independent, as the loss function and evaluation metric. 

\subsubsection{Results for \Algo (Figure~\ref{fig:pde}, left)}
Unlike the baseline methods which are trained autoregressively, \Algo is trained with single-step prediction, i.e., we feed the data at the first time step to the network to predict that of the last time step (output of the solver). This significantly improves computational efficiency but also increases the learning difficulty. Yet \Algo is still able to achieve smaller nMSEs relative to the baselines on most datasets. We also report \Algo's training time (stage 1, 2, and 3 combined) in Table~\ref{table:timepde}, which shows that cross-modal transfer is often both faster and more effective than domain-specific models.

\begin{table}[h]
\vspace{0.3cm}
\caption{Normalized RMSEs ($\downarrow$) on 8 PDEBench datasets, with baseline results  taken from \citet{Takamoto2022PDEBENCHAE}. Note that we only evaluated  datasets that can fit into a single NVIDIA V100 GPU, and the U-Net results for Naiver-Stokes and Darcy Flow are missing becuase the benchmark paper does not evaluate them also dueto memory issues. On 4 of  8 datasets, \Algo achieves the lowest nRMSEs. This aggregate result is the best even when compared with highly specialized neural operators such as FNO.
		}
		\label{table:pdebench1}
		\centering
\resizebox{0.9\textwidth}{!}{		\begin{tabular}{clcccccc}
			\toprule
			Dimension & Dataset  &Resolution & Parameters & PINN & FNO &U-Net& \Algo    \\
			
			\midrule
	\multirow{5}{40pt}{1D}	& Advection  & 1024 &$\beta=0.4$&6.7E-1 	&1.1E-2	& 1.1 &	\textbf{9.8E-3}\\
  & Burgers &  1024  &$\nu=1.0$ &3.6E-1	& \textbf{3.1E-3}	& 9.9E-1&1.2E-2\\
   & Diffusion-Reaction  & 1024 & $\nu=0.5$, $\rho=1.0$&6.0E-3 	&\textbf{1.4E-3 }&	8.0E-2&3.0E-3\\
  & Diffusion-Sorption & 1024 & -& 1.5E-1&	 1.7E-3 &	2.2E-1&	\textbf{1.6E-3}\\
   & Navier-Stokes &  1024 & $\eta = \zeta=0.1$, rand\_periodic& 7.2E-1 &	6.8E-2&-& \textbf{6.2E-2}\\
   \midrule
  \multirow{3}{40pt}{2D} 
    &   Darcy Flow&  128$\times$128 & $\beta=0.1$ &1.8E-1 	&2.2E-1&-&\textbf{8.1E-2}\\
    &   Shallow-Water&  128$\times$128 & -&8.3E-2 	&\textbf{4.4E-3 }	&1.7E-2&6.0E-3	\\
     &  Diffusion-Reaction   & 128$\times$128 & -  &8.4E-1& \textbf{1.2E-1}	& 1.6	&8.2E-1\\
			\bottomrule
		\end{tabular}}

  \end{table}

\begin{table}[h]
\vspace{0.3cm}
 \caption{Per epoch and total training time   for each method evaluated in Table~\ref{table:pdebench1}. Baseline numbers are taken from \cite{Takamoto2022PDEBENCHAE}. On 1D tasks, though it takes longer time for \Algo-RoBERTa to iterate over the entire dataset, our method converges faster, so overall \Algo  is still  more efficient than FNO and U-Net.}
\label{table:timepde}
	\centering
\resizebox{1.0\textwidth}{!}{	\begin{tabular}{lc|ccc|ccc|ccc|ccc}
		\toprule
      &   && FNO & & &U-Net&&& PINN& & &\Algo &   \\ 
		   Task   & Resolution & Per epoch (s) &Epoch& Total  (hrs)& Per epoch (s) &Epoch& Total  (hrs)& Per epoch (s) &Epoch& Total  (hrs)& Per epoch (s) &Epoch& Total  (hrs) \\ 
		\midrule
		
   Diffusion-Sorption & $1024^1$  & 97.52& 500   &
13.5 & 96.75&500 &
13.4 &0.011 &15000&0.046 &149.57 &200&
8.43 \\
      Shallow-Water& $128^2$ & 105.16&500&
14.6  & 83.32&500&11.6
& 0.041 &15000&0.17& 35.5 &200&
2.2\\
		\bottomrule
	\end{tabular}}

\end{table}

\subsubsection{Results for Zero-Shot Super-Resolution (Figure~\ref{fig:pde}, right)}
In addition to the above experiments, we also study whether under certain conditions, \Algo can achieve zero-shot super-resolution as described in \citet{li2021fno}. We see that when using convolution with kernel size 1 and the RoBERTa backbone, \Algo can indeed generalize to higher-resolution inputs. The detailed results are as follows.

\begin{table}[h]
		\vspace{0.3cm}
    \caption{We study   zero-shot super-resolution (trained on lower resolution and tested on higher resolution) on  the 1D Advection problem. \Algo-RoBERTa   achieves this  since the nRMSEs are similar across rows for different train-test resolution pairs. Note that the metrics differ slightly from the one reported in Table~\ref{table:pdebench1} because the kernel size of the convolution layer in  the embedder is searched via ASHA for experiments in Table~\ref{table:pdebench1}, whereas pointwise convolution with kernel size 1 is used to achieve super-resolution for experiments in this table.
		}
		\label{table:pdebench2}
  \centering
\begin{tabular}{lccc}
			\toprule
			  & Train Resolution (Spatial)  & Test Resolution (Spatial)& nRMSE   \\
			
			\midrule
			1D Advection  & 256&
			256& \res{1.13E-2}{2.71E-4}
\\
   1D Advection  & 256&
			512& \res{1.27E-2}{9.54E-5}
\\
   1D Advection  & 512&
			512&\res{1.02E-2}{2.37E-4}
\\
			\bottomrule
		\end{tabular}

\end{table}

\subsubsection{Results for fine-tuning and train-from-scratch baselines}
Similar to the NAS-Bench-360 experiments, we also want to study how much data alignment and knowledge transfer from pretrained models benefit downstream adaptation for PDE tasks. Therefore, we compare \Algo with the vanilla fine-tuning baseline (without data alignment) and the train-from-scratch baseline. As shown in the table below, these two baselines underperform ORCA, which shows the importance of distribution alignment. Besides, fine-tuning outperforms train-from-scratch on 5/8 tasks. This shows that whether transferring pretrained knowledge can benefit downstream adaptation is task-dependent. In some cases, naive fine-tuning without data alignment can even harm transfer.

\begin{table*}[h]
		\vspace{0.3cm}
  \caption{Normalized RMSEs ($\downarrow$) with error bars of \Algo, vanilla fine-tuning, and training RoBERTa/Swin from scratch on PDEBench datasets.
		}
		\large
\resizebox{1.0\textwidth}{!}{		\begin{tabular}{lcccccccc}
			\toprule
			 & Advection &  Burgers& Diffusion-Reaction& Diffusion-Sorption& Navier-Stokes& Darcy Flow& Shallow-Water& Diffusion-Reaction\\	
			\midrule
			Train-from-scratch & \res{1.7E-2}{7.0E-4}& \res{1.3E-2}{4.6E-4}& \res{1.7E-2}{2.2E-4}& \res{3.2E-3}{1.0E-6}& \res{9.9E-1}{3.6E-6}& \res{9.0E-2}{3.6E-3}& \textbf{\res{6.0E-3}{3.5E-6}}& \res{8.4E-1}{1.8E-3}\\
				Fine-tuning & \res{1.4E-2}{1.7E-3}& \res{1.4E-2}{3.6E-4}& \res{9.3E-3}{5.7E-3} & \res{3.1E-3}{6.5E-5}& \res{9.9E-1}{2.0E-5}& \textbf{\res{8.1E-2}{2.5E-3}}& \res{6.1E-3}{7.3E-6}& \res{8.3E-1}{9.3E-5}\\
		\Algo & \textbf{\res{9.8E-3}{1.4E-4}} & \textbf{\res{1.2E-2}{3.6E-4}}& \textbf{\res{3.0E-3}{1.5E-4}}& \textbf{\res{1.6E-3}{1.7E-4}}&\textbf{ \res{6.2E-2}{1.9E-3}}& \textbf{\res{8.1E-2}{8.1E-4}}& \textbf{\res{6.0E-3}{4.5E-6}}& \textbf{\res{8.2E-1}{4.6E-5}}\\
			
			\bottomrule
		\end{tabular}}
  
	\end{table*}

\subsection{Experiments on OpenML Tabular Datasets}
\label{appendix:openml}
We obtain the datasets using the built-in {get\_dataset} function of the \href{}{openml library}. For preprocessing, we follow the procedure in \citet{Dinh2022LIFTLF}. Specifically, we first remove all the rows whose labels are NaN and  drop the columns with missing entries. Then, we normalize the columns as follows:
\begin{itemize}
    \item Numerical features: we use the StandardScaler  class in sklearn to scale the data to have zero mean and unit variance and then concatenate all numerical features as one feature
    \item Categorical features: one-hot encoding is used
\end{itemize}
For training, we use the cross-entropy loss as the loss function, with the class weights set to $1/ (num\_class\_samples)$. 

\subsubsection{Complete Results for Table~\ref{table:tabsummary} (Top)}
To compare with TabPFN \citep{Hollmann2022TabPFNAT} and use the baselines reported in their paper,  we follow the same evaluation protocol and use the OVO (one-vs-one) AUROC (Area Under the ROC curve) as the score metric. The train-test split ratio is 0.5:0.5 to account for the limited context length of TabPFN.
The detailed results for each method on each task is shown in Table~\ref{table:tabular_results_table}, with the task meta-data shown in Table~\ref{table:metadata1}. We can see that there is 
not a single classification method that performs best on \textit{all}  datasets. However, \Algo obtains  good aggregate results in general, and its good performance on many challenging datasets where other baselines do no perform well makes it quite useful in real-life scenarios.

We also report the training time for each method in Table~\ref{table:metadata1}, which shows that \Algo does not take significantly longer time than  non-deep-learning-based methods. We emphasize that our method needs to be trained on a per-task basis. This is in contrast with TabPFN, which first fits a general prior network offline and then  for every new task, inference can be performed online within seconds. 

Besides, it is worth noting that one concern with using pretrained language models to solve tabular tasks is that these models might have seen the tabular data during pretraining. This may affect the test metrics, but we currently do not have a method to verify the degree of the effect.

 \begin{table}[h]
\caption{One-vs-one AUROC ($\uparrow$) on  30 OpenML-CC18 datasets. Baseline numbers are taken from \cite{Hollmann2022TabPFNAT}. \Algo achieves the best overall performance. }
    \label{table:tabular_results_table}
    \centering
\resizebox{1.0\textwidth}{!}{\begin{tabular}{l|ccc|cc|c}
\toprule
&        LightGBM &        CatBoost &          XGBoost &                      AutoGluon &            TabPFN  &\Algo-RoBERTa\\
\midrule
balance-scale        &          0.9938 &          0.9245 &           0.9939 &                  0.9919   & \textbf{ 0.9973 }  &0.9949
\\
mfeat-fourier        &          0.9786 &          0.9816 &           0.9803  &         \textbf{0.9843}   &           0.9811  &0.9729
 \\
breast-w             &           0.991 &          0.9931 &           0.9896  &                  0.9933 &         0.9934  &\textbf{0.9939}
\\
mfeat-karhunen       &          0.9979 &          0.9986 &           0.9983 &      \textbf{   0.9987 } &           0.9978   &0.9968
\\
mfeat-morphologica.. &          0.9601 &          0.9629 &           0.9612 &                  \textbf{0.9698}  &           0.9669 &0.9647  \\
mfeat-zernike        &          0.9716 &          0.9759 &           0.9735 &         \textbf{0.9908} &           0.9823   &0.9829
\\
cmc                  &          0.7288 &          0.7256 &          \textbf{ 0.7299 }&                  0.7331  &           0.7276 &0.7237
 \\
credit-approval      &          0.9415 &          0.9389 &  \textbf{0.9422}  &                  0.9415&           0.9322 &0.934
\\
credit-g             &          0.7684 &          0.7852 &           0.7853  &                  \textbf{0.7941}   &           0.7894  &0.7748
 \\
diabetes             &          0.8247 &          0.8383 &           0.8378  &                  0.8391   &           \textbf{ 0.841}  &0.8239 \\
tic-tac-toe          &          0.9988 &          0.9992 &       \textbf{1}  &              \textbf{1}   &           0.9759 &0.9973
 \\
vehicle              &          0.9232 &          0.9302 &           0.9282  &                  0.9416   &  0.9589  &\textbf{0.9591}
\\
eucalyptus           &          0.8931 &          0.8979 &           0.9004 &                  0.9204   &           \textbf{0.9245} &0.9084
 \\
analcatdata\_author.. &          0.9999 &          0.9999 &           0.9997   &                  0.9993   &       \textbf{1}  &0.9996
 \\
analcatdata\_dmft     &          0.5461 &          0.5589 &           0.5743  &                  0.5657  &  \textbf{ 0.579} &0.5627
 \\
pc4                  &          0.9301 &          0.9413 &           0.9291   &                \textbf{  0.9428}   &           0.9383  &0.9226 \\
pc3                  &          0.8178 &          0.8247 &           0.8288   &                  0.8282   &  0.8373  &\textbf{0.8411}
\\
kc2                  &          0.8141 &          0.8323 &           0.8227   &                  0.8242 &  0.8346  &\textbf{0.8431}
\\
pc1                  &          0.8321 &            0.86 &           0.8489 &                  0.8578  &           0.8761   &\textbf{0.8767}\\
banknote-authentic.. &      \textbf{1} &      \textbf{1} &       \textbf{1}   &              \textbf{1}  &       \textbf{1}  &\textbf{1}
 \\
blood-transfusion-.. &          0.7144 &          0.7403 &           0.7312  &                  0.7364  &  0.7549 &\textbf{0.7565}
\\
ilpd                 &          0.6917 &          0.7279 &           0.7171   &                   0.723   &           0.7379  &\textbf{0.7419}
 \\
qsar-biodeg          &          0.9126 &          0.9217 &           0.9191   &                  0.9276  &         {  0.9336} &\textbf{0.9349}
 \\
wdbc                 &          0.9904 &          0.9931 &           0.9904   &                  0.9956  &  \textbf{0.9964} &0.9929 \\
cylinder-bands       &          0.8556 &          0.8757 &           0.8782   &      \textbf{   0.8878}  &           0.8336 &0.844
 \\
dresses-sales        &          0.5593 &          0.5696 &  0.5823 &                  0.5507  &           0.5376   &\textbf{0.6025}
 \\
MiceProtein          &          0.9997 &          0.9999 &           0.9998  &              \textbf{1}  &           0.9999 &0.9969
\\
car                  &          0.9925 &          0.9955 &           0.9948 &   0.998   &            0.995 &\textbf{0.9983}
 \\
steel-plates-fault.. &          0.9626 &          0.9655 &           0.9656  &                  \textbf{0.9666}   &           0.9655  &0.9543
\\
climate-model-simu.. &          0.9286 &          0.9344 &           0.9255  &                  0.9391  &           0.9415  &\textbf{0.9416}  \\
\midrule
\# Wins   &   1	&1	&3	&\textbf{12}	&7&	\textbf{12}  \\
\midrule
Avg. AUROC 
                                         &  0.884$\pm$0.1301 &   0.8898$\pm$0.1232 &   0.8909$\pm$0.1224 &              \textbf{0.8947$\pm$0.1266}  &    0.8943$\pm$0.1249&   0.8946$\pm$0.1206\\
\midrule
Avg. Diff. from XGBoost &  -6.97E-3$\pm$9.1E-3&   -1.18E-3$\pm$1.42E-2 &   0 &            \textbf{ 3.74E-3$\pm$9.18E-3}  &    3.38E-3$\pm$1.72E-2 &   3.63E-3$\pm$1.47E-2\\

\bottomrule
\end{tabular}}
\end{table}

\newpage
\begin{table}[h]
\caption{Meta-data for the OpenML-CC18 datasets taken from \citet{Hollmann2022TabPFNAT}. \Algo's training time depends on the size of the dataset as well as the sequence length of the generated features (note that the latter is determined by the kernel size in the embedder layer, which is searched via hyperparameter tuning).  Average training time for \Algo is 4 min  per dataset.}
    \label{table:metadata1}
    \centering
\resizebox{0.96\textwidth}{!}{\begin{tabular}{@{\hskip 0mm}c@{\hskip 0mm}ccccccc}
\toprule
 OpenML ID &  Name & \#Feat. &  \#Cat. &  \#Inst. &  \#Class.  &  Minor. Class Size & \Algo train time (min) \\
\midrule
11 & balance-scale & 5 & 1 & 625 & 3 & 49 & 8.7
\\
14 & mfeat-fourier & 77 & 1 & 2000 & 10  & 200&10.49
 \\
15 & breast-w & 10 & 1 & 699 & 2 & 241&1.29
 \\
16 & mfeat-karhunen & 65 & 1 & 2000 & 10  & 200 &3.84
\\
18 & mfeat-morphological & 7 & 1 & 2000 & 10   & 200 &21.46  \\
22 & mfeat-zernike & 48 & 1 & 2000 & 10 & 200&4.55
 \\
23 & cmc & 10 & 8 & 1473 & 3   & 333 &2.27
 \\
29 & credit-approval & 16 &    10 & 690 & 2  & 307&1.34
\\
31 & credit-g & 21 &    14 & 1000 & 2   & 300  &1.82
 \\
37 & diabetes & 9 & 1 & 768 & 2  & 268&1.43
\\
50 & tic-tac-toe & 10 &    10 & 958 & 2  & 332&1.50 \\
 54 &vehicle & 19 & 1 & 846 & 4  & 199 &2.10
\\
188 & eucalyptus & 20 & 6 & 736 & 5   & 105&2.06
 \\
 458 &analcatdata\_auth... & 71 & 1 & 841 & 4   & 55&2.08
  \\
469 & analcatdata\_dmft & 5 & 5 & 797 & 6   & 123 &2.17
\\
1049 & pc4 & 38 & 1 & 1458 & 2   & 178 &2.28
 \\
1050  & pc3 & 38 & 1 & 1563 & 2  & 160 &1.96\\
1063 & kc2 & 22 & 1 & 522 & 2   & 107 &1.10
 \\
1068 & pc1 & 22 & 1 & 1109 & 2 & 77&1.68
 \\
1462 & banknote-authenti... & 5 & 1 & 1372 & 2   & 610 &2.32
 \\
1464 &blood-transfusion-... & 5 & 1 & 748 & 2  & 178 &1.46
  \\
1480 &ilpd & 11 & 2 & 583 & 2   & 167 &1.17\\
1494 &qsar-biodeg & 42 & 1 & 1055 & 2   & 356&11.06
 \\
1510 &wdbc & 31 & 1 & 569 & 2   & 212 &1.23
 \\
6332 &cylinder-bands & 40 &    22 & 540 & 2  & 228&1.07
\\
23381 &dresses-sales & 13 &    12 & 500 & 2   & 210&1.47
 \\
40966 &MiceProtein & 82 & 5 & 1080 & 8   & 105 &2.51
 \\
40975 &car & 7 & 7 & 1728 & 4  & 65  &17.19
\\
40982 &steel-plates-fault & 28 & 1 & 1941 & 7   & 55 &5.83
 \\
40994 &climate-model-simu... & 21 & 1 & 540 & 2  & 46 &1.00 \\
\bottomrule
\end{tabular}}

\end{table}

\newpage
\subsubsection{Results for train-from-scratch and fine-tuning baselines on OpenML-CC18}

We run the fine-tuning and train-from-scratch baselines using the  train-test split scheme in \citet{Hollmann2022TabPFNAT} and compare their performance with \Algo. Unlike on NAS-Bench-360 and PDEBench, train-from-scratch performs better than fine-tuning on tabular tasks. This shows that initializing the network with out-of-modality pretrained weights may lead to suboptimal performance, which is also observed in several recent work \cite{Kumar2022FineTuningCD, Lee2022SurgicalFI}.

\begin{table}[h]
 \caption{\Algo vs. train-from-scratch and fine-tuning on  tabular tasks evaluated in \citet{Hollmann2022TabPFNAT}.   ``Diff. from XGBoost" is the across-task average of per-task difference from XGBoost. }
 \vspace{-0.1cm}
	\centering
 \Large
\resizebox{0.49\textwidth}{!}{	\begin{tabular}{l|ccc}
		\toprule
 OpenML-CC18   &  Train-from-scratch & Fine-tuning & \Algo \\
		\midrule
\# Wins/Ties    &   11/30	&1/30&	\textbf{20/30}  \\
Avg. AUROC ($\uparrow$) &  0.8673 & 0.8661 &   \textbf{ 0.8946}\\
Diff. from XGBoost &   -2.4E-2 & -2.5E-2& \textbf{+3.63E-3}\\ 
		\bottomrule
	\end{tabular}}
	\end{table}

 \subsubsection{Complete Results for Table~\ref{table:tabsummary} (Bottom)}
To compare with LIFT \citep{Dinh2022LIFTLF} and use the baselines reported in their paper,  we follow the same evaluation protocol and use the classification accuracy as the score metric. 
The detailed results for each method on each task is shown in Table~\ref{tab:classification_accuracy}, with the task meta-data shown in Table~\ref{tab:clf_datasets}.
 \begin{table}[h]
\vspace{0.3cm}
  \caption{
   Accuracies ($\uparrow$) on the classification tasks evaluated in \cite{Dinh2022LIFTLF}. Baselines include LIFT, the prompting method which that uses large-scale pretrained language models, and standard ML methods such as XGBoost. \Algo achieves competitive performances with existing methods and ranks first on 7 out of 14 datasets, significantly outperforming the domain-specific cross-modal learning approach, LIFT. 
    }
\label{tab:classification_accuracy}
\centering
    \begin{adjustbox}{width=0.95\textwidth,center}
    \begin{tabular}{l|cccc|c|c}
    \toprule[0.05cm]
Dataset (ID)   &Logistic Regression   & Decision Tree    & SVM    & XGBoost & LIFT w. GPT-3 & \Algo \\[0.05cm] \midrule 
       Customers (1511)    & \textbf{\res{87.12}{0.54}} &  \res{85.98}{0.53} &  {\res{86.36}{0.00}} & \res{85.23}{0.00}   & \res{84.85}{1.42}&\res{86.93}{1.13}
 \\
  Pollution (882)     & \res{58.33}{11.79} &  \textbf{\res{77.78}{3.93}} &  \res{58.33}{6.81} &  \res{63.89}{7.86}   & \res{63.89}{7.86} &\res{75.00}{9.62}
\\
        Spambase (44)    & \res{93.27}{0.00} & \res{ 90.7}{0.14}  & \res{93.70}{0.00} &  \textbf{\res{95.87}{0.00}}    & \res{94.90}{0.36}&\res{94.36}{0.17}
 \\
   Hill-Valley (1479)  & {\res{77.78}{0.00}} &  \res{56.38}{0.89} & \res{68.72}{0.00} &  \res{59.26}{0.00}   & \textbf{\res{99.73}{0.19}}&\res{74.86}{2.06} \\
  IRIS (61) &  \res{96.67}{0.00} & \res{97.77}{3.85}  &  \textbf{\res{100.00}{0.00}}  & \textbf{\res{100.00}{0.00}}    &   \res{97.0}{0.00} &\textbf{\res{100.00}{0.00}}\\
   TAE (48)   & \res{45.16}{4.56} & {\res{65.59}{5.49}}  & \res{53.76}{6.63} &  \res{66.67}{8.05}  & \res{65.59}{6.63}&\textbf{\res{70.31}{5.98} }
 \\
      CMC (23)    & \res{49.49}{0.83} &  {\res{56.72}{0.32}} &  \res{56.50}{0.97} &  \res{52.43}{0.42}  & \res{57.74}{0.89} &\textbf{\res{58.11 }{ 1.78} }
 \\
    Wine (187)   & \textbf{\res{100.00}{0.00}}  & \res{93.52}{2.62} &  \textbf{\res{100.00}{0.00}} &  \res{97.22}{0.00}  & \res{92.59}{1.31}&\res{98.61}{2.77}  \\
  Vehicle (54)   & {\res{80.39}{1.00}} &  \res{63.92}{2.37} &  \res{81.18}{0.48} &  \res{73.14}{0.28}  & \res{70.20}{2.73} &\textbf{\res{82.35}{0.96}} 
\\
     LED (40496)  & \res{68.67}{0.94} &  \res{66.33}{2.87} &  \res{68.00}{0.82} &  \res{66.00}{0.82}  & {\res{69.33}{2.05}}&\textbf{\res{71.50}{2.51}}
 \\
   OPT (28)     & \res{96.53}{0.22} & \res{89.8}{1.09} &  {\res{97.95}{0.00}} &  \res{97.48}{0.17}   & \textbf{\res{98.99}{0.30}} &\res{98.09}{0.39}\\
   Mfeat (12)       & \res{97.67}{0.12} &  \res{87.67}{1.05} &  \textbf{\res{98.83}{0.24}} &  \res{96.75}{0.00}   & \res{93.08}{0.24}&\res{96.88}{1.03} \\
 Margin (1491)   & {\res{81.35}{0.15}} & \res{ 43.86}{1.21}  & {\res{81.98}{0.30}} & \res{70.21}{0.29}   & \res{59.37}{0.92} & \textbf{\res{82.65}{0.59}}\\
  Texture (1493)    & {\res{81.67}{0.97}} &  \res{46.88}{1.93} & {\res{83.44}{0.89}} &  \res{70.73}{1.41}   & \res{67.50}{1.42} & \textbf{\res{83.59}{2.35}}\\
          \midrule
\# Wins/Ties   &  2&	1	&3&	2&	2&	\textbf{7} \\
\midrule
Avg. Acc
                                         & \res{79.58}{18.06}	& \res{73.06}{18.17} 	& \res{80.63}{16.87}	& \res{78.21}{16.57}	& \res{79.63}{16.09}	& \textbf{\res{83.80}{12.81}}\\
    \midrule
    Avg. Diff. from XGBoost & \res{1.37}{9.42}&	\res{-5.14}{10.33}&	\res{2.42}{6.84} & 0 &	\res{1.42}{11.88}&	\textbf{\res{5.60}{5.66}}\\
    \bottomrule[0.05cm]
    \end{tabular}
    \end{adjustbox}
\end{table}%
\begin{table}[h] 
\vspace{0.3cm}
   \caption{Meta-data for the OpenML classification datasets evaluted in Table~\ref{tab:classification_accuracy}. Taken from \citet{Dinh2022LIFTLF}. }
    \label{tab:clf_datasets}
    \centering
    \scriptsize
\resizebox{0.9\textwidth}{!}{
    \begin{tabular}{lccccc}
        \toprule 
    ID &  {Abbreviation} & { No. Features } &  {No. Classes } &   {No. Instances }  &  { Note } \\ 
        \midrule
    1511 & Customers  & 8& 2 & 440& Imbalance \\
   882 &  Pollution & 15 & 2 & 60&  1 symbolic feature\\
         44 & Spambase & 57 & 2 & 4601 & 1 symbolic feature\\
      1479 & Hill-Valley & 100 & 2 & 1212 & 1 symbolic feature\\
      48 & TAE & 5 & 3 & 151 & Categorical data \\
     23 & CMC & 9& 3& 1473& Meaningful feature Names\\
      187 & Wine & 13& 3&178 & Integral features\\ 
         54 & Vehicle & 18& 4& 846& Meaningful feature Names\\ 
        40496& LED & 7 & 10 & 500 & 1 symbolic feature \\
      28 & OPT&64 & 10 & 5620 & 1 symbolic feature\\
     12 & Mfeat &  216 & 10 & 2000& 1 symbolic feature\\
    871 & Pollen & 5 & 2 & 3848 & - \\
     1467 & Climate & 20 & 2 & 540 & -\\
      1491& Margin & 64 & 100 & 1600 & 1 symbolic feature\\
        1492 & Shape & 64& 100 & 1600& 1 symbolic feature\\
      1493 & Texture& 64& 100 & 1599& 1 symbolic feature\\
        \bottomrule
    \end{tabular}}
  
\end{table}

\newpage
\subsection{Experiments on Drug Response Prediction}
\label{appendix:drug}
We adapt the code from the \href{https://github.com/zhuyitan/IGTD}{official GitHub repo} of IGTD and download, preprocess, and load the CTRP \& GDSC data following the procedures described in the paper's \href{https://static-content.springer.com/esm/art%3A10.1038%2Fs41598-021-90923-y/MediaObjects/41598_2021_90923_MOESM1_ESM.pdf}{supplementary material}. Notably, both the gene expression data and the drug descriptors are normalized using min-max normalization so that each gene/drug feature has a maximum value of 1 and a minimum value of 0. We then concatenate the features for each gene-drug (treatment) pair. The number of features (cols) for each treatment sample (row) is 3901 for CTRP and 3739 for GDSC. The processed data are stored locally for the ease of data loading. 
During training, we use the MSE as the loss function since we are in a regression setting. The prediction performance is measured by the coefficient of determination ($R^2$).

Table~\ref{table:drug}, we show the results for \Algo and the baselines, which include the domain-specific IGTD algorithm that transforms  gene expression profiles and drug molecular descriptors into their respective images. We can see that even compared with such highly specialized algorithms, the domain-agnostic \Algo still performs quite well, showing the capacity of cross-modal transfer learning with  large-scale pretrained models.

\subsection{Additional Experiments}
\subsubsection{Compatibility with In-Modality Transfer}
\label{appendix:inmodal}
\begin{wraptable}{r}{0.5\textwidth}
  	\caption{\small  We use the dataset splits in  
		\citet{Tan2020ClassImbalancedDA}, which removed some mislabeled outliers, and report the prediction errors ($\downarrow$) for  \Algo and fine-tuning (using Swin-base).
		 }
		\label{table:domainnet}
	
		\centering
		\small
		\resizebox{0.5\textwidth}{!}{\begin{tabular}{lcccc}
			\toprule
			     &  Real  &  Painting & Sketch & Clipart  \\ 
			\midrule
		\Algo   & \textbf{\res{96.71}{0.02}}&\textbf{\res{94.71}{0.13}}&\textbf{\res{94.93}{0.24}}  &\textbf{\res{93.61}{0.54}} \\
			Fine-tuning   &\res{93.33}{1.33}&\res{75.79}{0.86}&\res{83.00}{0.13} & \res{86.01}{2.62} \\
			\bottomrule
		\end{tabular}}

  \vspace{5mm}
	\end{wraptable}
A natural question to ask is whether \Algo can also  tackle in-modality tasks. While we design \Algo to enable cross-modal transfer, we hypothesize that it should facilitate same-modality transfer if two domains have large dataset distance. To validate this, we test \Algo on DomainNet datasets, which are commonly used to evaluate homogeneous DA methods \citep{peng2019moment}. From Table~\ref{table:domainnet}, we can see that \Algo achieves significantly better performance than the fine-tuning baseline, which shows that the feature matching of \Algo can also help in-domain generalization.

\subsubsection{Prompting}	
\label{appendix:prompting}
Apart from fine-tuning, a new paradigm of working with large-scale pretrained models is prompting, i.e., we do not update the pretrained weights but only modify the input and query the model for the desired output. Existing language prompting methods \citeg{Liu2022PretrainPA} are generally not suitable for cross-modal learning due to the difficulty of designing natural prompts for diverse data types. For the 1D tasks we study, there is even no notion of ``discrete tokens." 
Another line of work studies visual prompting by modifying 2D inputs for querying vision transformers. We test two such algorithms, VP \citep{Bahng2022ExploringVP} and VPT \citep{Jia2022VisualPT}, on three classification tasks in our task suite. They are not applicable to the remaining tasks because either the inputs cannot be reshaped to look like images or the outputs are not classification logits. 

\begin{wraptable}{r}{0.53\textwidth}
\caption{\small Prediction errors ($\downarrow$) of \Algo vs. visual prompting methods.   
		}
		\vspace{-0.1cm}
		\label{table:prompting}
\centering
\vspace{-0.1cm}
		
		\begin{tabular}{lccc}
			\toprule
			  &Spherical&NinaPro & ECG\\	
			 \midrule
\Algo&\textbf{\res{29.85}{0.72}} &\textbf{ \res{7.54}{0.39}} & \textbf{\res{0.28}{0.0059}  }\\
VP &\res{98.05}{0.13} & \res{33.18}{0.23}&\res{0.57}{0.0044}\\
VPT &\res{49.53}{1.45} & \res{31.46}{0.83} & \res{0.40}{0.016}  \\
			\bottomrule
		\end{tabular}

	\end{wraptable}
We test VPT with the pretrained Swin-Base Transformer (the same model we used for \Algo) and VP with the pretrained ResNet-50 (as the official implementation does not support vision transformers). The results are shown in Table~\ref{table:prompting}. In general, prompt tuning is less effective than fine-tuning, and the two baselines perform significantly worse than \Algo. This is not surprising given that prompting methods are more intuitively suited to in-modality transfer, where the target and the source data have similar structure or semantic meaning. However, when the target data (e.g., electromyography signals, as in the NinaPro dataset) is drastically different from image data, it is difficult to design prompts or expect good performance by only modifying the inputs without fine-tuning the pretrained models.


\end{document}